\documentclass[10pt,twocolumn,letterpaper]{article}

\usepackage[pagenumbers]{cvpr} %

\usepackage{graphicx}
\usepackage{amsmath}
\usepackage{amssymb}
\usepackage{booktabs}

\usepackage[export]{adjustbox}

\makeatletter
\@namedef{ver@everyshi.sty}{}
\makeatother

\usepackage{booktabs}
\usepackage{tikz}
\usetikzlibrary{calc}

\usepackage{float}
\usepackage{longtable}

\usepackage{xcolor}

\definecolor{mlblue}{rgb}{0 0.4470 0.7410}
\definecolor{mlgreen}{rgb}{0.4660 0.6740 0.1880}
\definecolor{mlred}{rgb}{0.6350 0.0780 0.1840}

\newcommand{\vect}[1]{\mathbf{#1}} %
\newcommand{\mat}[1]{#1}%
\newcommand{\surf}[1]{\mathcal{#1}} %

\newcommand{\ppoint}{\vect{p}}
\newcommand{\qpoint}{\vect{q}}

\usepackage[pagebackref,breaklinks,colorlinks]{hyperref}

\usepackage[capitalize]{cleveref}
\crefname{section}{Sec.}{Secs.}
\Crefname{section}{Section}{Sections}
\Crefname{table}{Table}{Tables}
\crefname{table}{Tab.}{Tabs.}

\def\cvprPaperID{3512} %
\def\confName{CVPR}
\def\confYear{2022}

\begin{document}

\title{Gradient-SDF: A Semi-Implicit Surface Representation for 3D Reconstruction}

\author{Christiane Sommer$^{*}$ \quad Lu Sang$^{*}$ \quad David Schubert \quad Daniel Cremers\\
Technical University of Munich\\
Computer Vision Group\\
{\tt\small \{sommerc, sang, david.schubert, cremers\}@in.tum.de}
}

\maketitle

\let\thefootnote\relax\footnote{$^{*}$ These authors contributed equally.}

\begin{abstract}
We present \emph{Gradient-SDF}, a novel representation for 3D geometry that combines the advantages of implict and explicit representations.
By storing at every voxel both the signed distance field as well as its gradient vector field, we enhance the capability of implicit representations with approaches originally formulated for explicit surfaces.
As concrete examples, we show that (1) the Gradient-SDF allows us to perform direct SDF tracking from depth images, using efficient storage schemes like hash maps, and that (2) the Gradient-SDF representation enables us to perform photometric bundle adjustment directly in a voxel representation (without transforming into a point cloud or mesh), naturally a fully implicit optimization of geometry and camera poses and easy geometry upsampling.
Experimental results confirm that this leads to significantly sharper reconstructions.
Since the overall SDF voxel structure is still respected, the proposed Gradient-SDF is equally suited for (GPU) parallelization as related approaches.
\end{abstract}

\section{Introduction}

\begin{figure}[t]
    \centering
\hfill
\includegraphics[width=.65\linewidth, angle=-90]{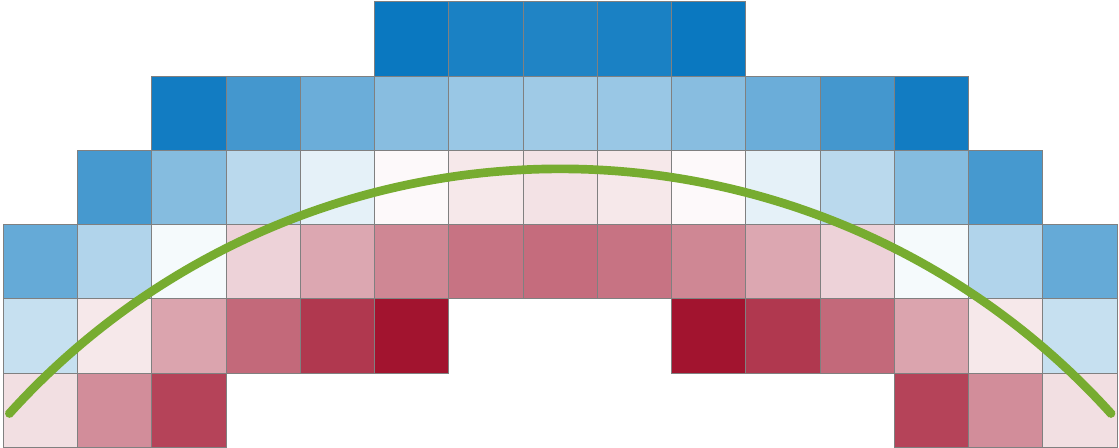}
\hfill\hfill
\includegraphics[width=.65\linewidth, angle=-90]{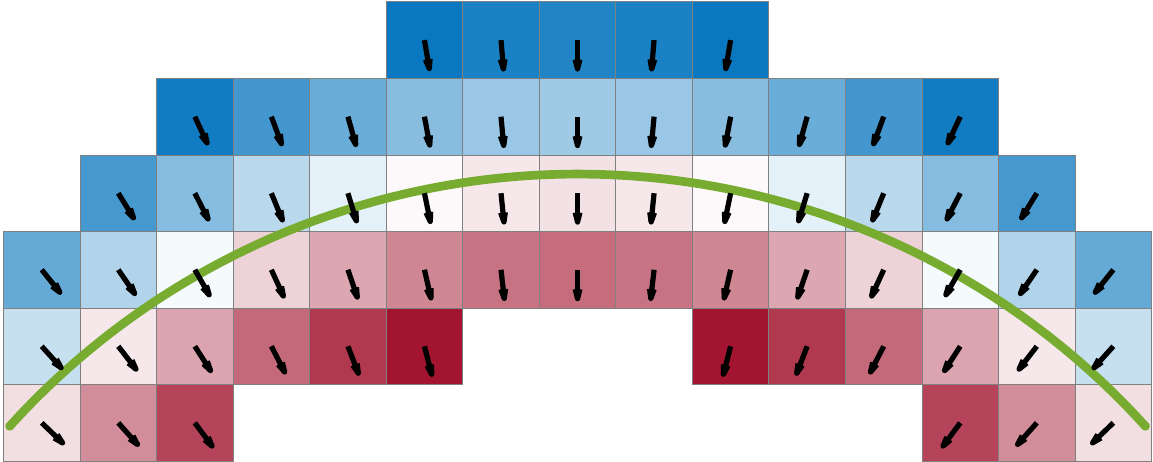}
\hfill\hfill
\includegraphics[width=.65\linewidth, angle=-90]{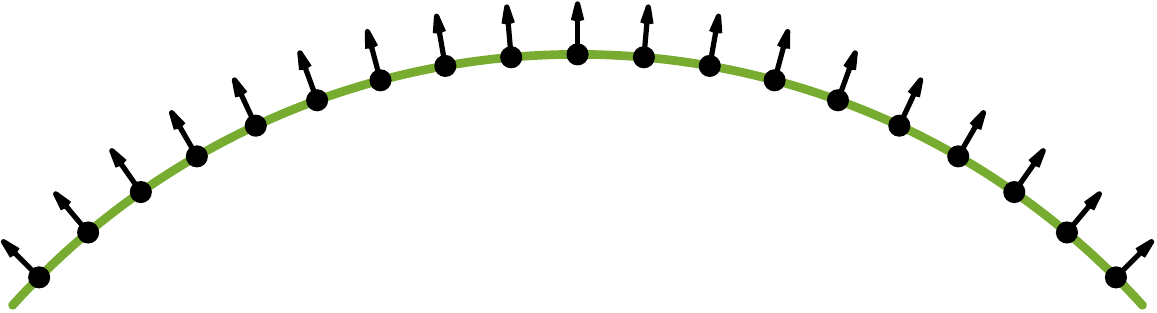}
\hfill{\color{white}.}
\\ \vspace{1.5mm}
\caption{Our Gradient-SDF (\emph{middle}) is a hybrid representation between standard signed distance fields stored in a voxel grid (\emph{left}) and the explicit geometry representation using surfels (\emph{right}): while we inherit the implicit nature of standard SDF voxels, we store gradients per voxel, which is similar to the surface normal property of a surfel. This combines the advantages of implicit representations, such as the possibility for direct SDF tracking, with those of explicit ones, for instance the possibility to perform bundle adjustment.}
\label{fig:surfels_voxels}
\end{figure}

The representation of 3D geometry in computer vision is a long-studied research topic.
Mathematically, a surface is a 2D manifold embedded in $\mathbb{R}^3$.
However, when it comes to implementing this on a computer, the question of discretization comes up:
how can we represent a surface with possibly infinite amount of detail and large extent with a finite amount of memory and variables with finite precision?
Different answers to this question exist, and which one is most suitable highly depends on the concrete problem one wants to solve.

On the one hand, there are \emph{explicit representations}, such as point clouds, surfel clouds or polygon meshes.
They directly sample points on the surface together with additional information like surface normals or point radius (for surfels), or connectivity of points (for meshes).
This is useful for applications such as bundle adjustment, where surface points are reprojected into different camera frames.

On the other hand, we have \emph{implicit representations}, that take a different approach:
the surface is encoded implicitly by assigning each point $\ppoint$ in the ambient space $\mathbb{R}^3$ a scalar value, such as a binary occupancy, or a (signed) distance to the nearest surface point.
\emph{Signed distance fields} (SDFs) have some useful properties, for instance, unlike explicit representations, they allow for changes of surface topology, and they can be updated very easily.
There are different ways to store SDFs, the more traditional one being a voxel grid, where 3D space is partitioned in voxels, \ie, cubes of a given size.
Each voxel contains the SDF value of its center point, sometimes truncated to a certain value.
Such voxel grids can be stored either densely, or sparsely using \eg octrees or hash maps.
Extracting a surface from an implicit representation requires an additional step, marching cubes being a popular choice~\cite{lorensen1987marching}.
Only recently, implicit parameterizations of implicit representations have become more popular.
Neural networks regress the SDF value for a surface at any point, and the set of network weights uniquely characterizes the surface geometry.
While these \emph{doubly implicit} representations are very elegant as they theoretically provide an infinite level of detail, updating the geometry is less trivial than for voxel-based implicit representations, and surface extraction often requires voxelization.

In this work, we propose a hybrid between explicit and implicit representations called \emph{Gradient-SDF}:
we use a voxel-based implicit SDF representation, and augment it with the SDF gradient.
To summarize, we propose the following contributions:
\begin{itemize}
    \item We propose Gradient-SDF as an implicit geometry representation with explicit features. It exploits first order Taylor expansion to perform interpolation without accessing several voxels.
    \item We prove that our stored Gradient-SDF vectors are significantly more accurate than gradients obtained by standard finite difference schemes.
    \item We show theoretically and experimentally how Gradient-SDF can be used in a depth-based tracking and mapping system, where efficient storage in a hash map is combined with direct SDF tracking.
    \item We provide a formulation for photometric bundle adjustment (BA) on our implicit voxel-based Gradient-SDF representation and evaluate the benefits of this. %
\end{itemize}

\section{Related Work}

\paragraph{ICP tracking and SDF-based mapping}

In KinectFusion~\cite{newcombe2011kinectfusion}, the depth camera is tracked by the iterative closest point (ICP) algorithm, where the incoming depth map point cloud is registered to a point cloud extracted from the current SDF model via raycasting.
Many methods build on this idea and focus on improving different aspects, most notably on reducing the high memory requirements imposed by the need for a volumetric voxel grid.
The voxel hashing approach set the basis for lower memory requirements~\cite{niessner2013real}, and others followed and further improved this~\cite{kahler2015very, whelan2015real, kahler2016hierarchical}.
In TextureFusion~\cite{lee2020texturefusion}, the voxel hashing data structure is augmented by a texture tile in order to compute high-resolution texture within an RGB-D scanning setup.
While these KinectFusion-like approaches work very well and can be efficiently implemented on a GPU, they switch between different representations of 3D geometry:
the voxel grid stores the SDF and is used for mapping, and a 3D point cloud obtained by raycasting is used for tracking.

\paragraph{SDF-based tracking and mapping}

Few works have addressed this issue of representation switching, most notably Bylow \etal~\cite{bylow2013real} and Canelhas \etal~\cite{canelhas2013sdf}, who directly minimize a sum of squared SDFs to estimate the camera pose, rather than converting the SDF to a point cloud.
Slavcheva \etal~\cite{slavcheva2016sdf} also convert the input depth map to a volumetric SDF prior to camera pose estimation.
Both types of approaches converge better than KinectFusion-like methods, while avoiding the change of representation.
However, they heavily rely on the voxel being stored cohesively in memory, as the tracking step involves interpolation and gradient computation of the SDF, which needs to access at least eight voxels for one single lookup.

\paragraph{Surfel-based tracking and mapping}

Surfel-based methods can also achieve impressive results for 3D tracking and mapping from depth images.
Keller \etal~\cite{keller2013real} use surfels, \ie, points together with their normals and some additional properties, to reconstruct 3D scenes from depth sensors.
To find surfels that correspond to pixels in an incoming image, they use index maps and align the incoming depth map with a virtual model depth map using ICP.
One particular advantage of explicit representations is the fact that dense photometric bundle adjustment~\cite{delaunoy2014photometric} can be easily integrated into a tracking and mapping pipeline, a recent example being BAD SLAM~\cite{schops2019bad}.
Thanks to SurfelMeshing~\cite{schops2019surfelmeshing}, it is even possible to extract meshes from a surfel representation, using a lazily updated octree to store and easily access surfels.

\paragraph{Neural networks for tracking and mapping}

In the last years, neural networks have been used increasingly for tracking and mapping, as the representation of geometry using learned parameters has proven very successful~\cite{mescheder2019occupancy,park2019deepsdf,gropp2020implicit,ma2021neural}.
In RoutedFusion~\cite{weder2020routedfusion}, the SDF is still stored as discrete voxel grid values, but the SDF update for a new incoming depth image is learned.
The later NeuralFusion work~\cite{weder2021neuralfusion} also represents geometry in a latent space.
Similar approaches are also taken in more recent work~\cite{azinovic2021neural,sucar2021imap}.

\section{Geometry Representations for 3D Vision}

\subsection{Surfels or voxels?}
\label{sec:mapping}

\emph{SDFs defined on a voxel grid} have been used for decades to store and efficiently update 3D geometry~\cite{curless1996volumetric}.
A subset $\Omega$ of $\mathbb{R}^3$ is subdivided into a discrete set of voxels with positions $\vect{v}_j\in\mathbb{R}^3$, each of which stores the distance $\psi_j$ to its closest surface point, see also Figure~\ref{fig:surfels_voxels} (\emph{left}).
The sign indicates if the voxel location is free space ($\psi_j<0$) or inside an object ($\psi_j>0$).
Such a free-space classification is useful for collision avoidance in navigation tasks.
Given an SDF representation and a new distance estimate $d(\vect{v}_j)$ for a voxel located at $\vect{v}_j$, $\psi_j$ is easily updated:
\begin{align}
    \label{eq:sdf_update}
    \psi_j &\leftarrow \frac{W_j\psi_j + w(\vect{v}_j)d(\vect{v}_j)}{W_j + w(\vect{v}_j)}\,, \\
    W_j &\leftarrow W_j + w(\vect{v}_j)\,,
\end{align}
with $w(\vect{v}_j)$ a weight indicating how reliable the estimate is, and $W_j$ the current weight estimate.
The set of voxels $\vect{v}_j$ is usually arranged on a regular 3D grid, $\vect{v}_j = v_s \vect{m}_j$ with voxel size $v_s$ and $\vect{m}_j\in\mathbb{Z}^3$.
If stored volumetrically, $\psi_j$ and $W_j$ can be accessed very quickly, but memory grows cubically with the scene size.
The memory footprint can be significantly reduced using hierarchical tree structures~\cite{steinbrucker2013large, steinbrucker2014volumetric} or hash maps~\cite{niessner2013real}.
For those, however, certain operations in the SDF, such as tri-linear interpolation or gradient computation (\ie, normal estimation) can become costly, as multiple voxels need to be accessed.
This is an issue, even if voxel blocks as in~\cite{niessner2013real} are used:
for a voxel block of size $8^3$, only $6^3$ voxels lie fully inside the block.
For these, neighbors can be accessed similarly to a volumetrically stored SDF.
However, for the $57.8\%$ of voxels ($8^3-6^3$) in the block that have neighbors outside their block, neighbor lookup still means additional hash table lookups.
Furthermore, to fully exploit the regular structure inside blocks, we would have to introduce a distinction between voxel types (fully inside/face/edge/corner).
In addition, in hash maps it is possible that not all eight voxel corners exist, meaning we need rules to perform the interpolation given only a subset of neighbors.

On the other side of the spectrum, there are \emph{surfels}:
the surface is explicitly represented by a set of points $\vect{p}$ on the surface, together with surface normals $\vect{n}$, see Figure~\ref{fig:surfels_voxels} (\emph{right}).
Surfels may have more properties, such as a radius, some visual descriptor, or timestamps~\cite{schops2019bad}, that can be used to define update rules in specific tracking or mapping applications.
Surfel location and normal can be updated similarly to \eqref{eq:sdf_update} in a running weighted average fashion.
Normal estimation in a surfel representation is trivial, as normals are stored inside the surfel data structure.
However, computing the distance to the closest surface point for a given point $\ppoint\in\mathbb{R}^3$ can become quite costly, as
it typically involves a nearest-neighbor search.
Furthermore, surfel representations don't have an easy way to classify non-surface points in $\mathbb{R}^3$, so they cannot be used for tasks in which we want to distinguish between free and occupied space.

\subsection{Gradient-SDF: the best of both worlds}

We propose \emph{Gradient-SDF} as a hybrid solution that combines the best of the implicit voxel world and the explicit surfel world:
we augment the voxel structure by an additional 3D vector, namely a scaled gradient $\vect{g}_j$ of the SDF at that point. %

This proposed data structure is visualized in Figure~\ref{fig:surfels_voxels}.
For a signed distance function, the gradient at a point $\ppoint$ is equal to the inwards-pointing surface normal at the closest surface point, and the negative of the outwards-pointing surface normal.
Thus, similarly to the update in \eqref{eq:sdf_update}, $\vect{g}_j$ can be updated in a straightforward way:
\begin{equation}
    \label{eq:gradient_update}
    \vect{g}_j \leftarrow \vect{g}_j + w(\vect{v}_j)\vect{n}(\vect{v}_j)\,.
\end{equation}
In most applications, normals are already computed from the incoming data (\eg, depth maps) for filtering or rendering, so the computation of $\vect{n}(\vect{v}_j)$ does not introduce any computational overhead.
We normalize the weighted sum $\vect{g}_j$ to get the actual gradient estimate $\hat{\vect{g}}_j$ at $\vect{v}_j$.

Storing the gradients together with the distances allows for easy computation of the closest surface point $\ppoint_s$ of a voxel $\vect{v}_j$:
\begin{equation}
\label{eq:grad_sdf_project}
    \ppoint_s(\vect{v}_j) = \vect{v}_j - \psi_j\hat{\vect{g}}_j\,.
\end{equation}
As each $\vect{v}_j$ can be uniquely mapped to a surfel with point $\ppoint_s(\vect{v}_j)$ and normal $-\hat{\vect{g}}_j$, we can interpret our storage scheme as a \emph{voxelized way to store surfels}.

Just like traditional voxel SDFs, our Gradient-SDF can be stored either volumetrically, or using optimized structures like trees or hash maps.
Since memory per voxel is increased by the use of Gradient-SDF, we focus our analysis on sparse voxel storage schemes.
Combining a voxel representation with components of a surfel representation (namely, the SDF gradient/surface normal), Gradient-SDF overcomes the issues that pure voxel or surfel representations may have, in particular when voxels are stored sparsely.
We demonstrate this on a range of example applications.

\section{Example Applications in 3D Vision}

\subsection{Camera tracking using depth images}

To find the rigid body transformation $(\mat{R},\vect{t})$ of an incoming point cloud with points $\ppoint_k$ to a global surface model $\surf{S}$, we aim to minimize a weighted least squares energy
\begin{equation}
\label{eq:tracking}
E(\mat{R},\vect{t}) = \sum_{k}{w_kd_\surf{S}(\mat{R}\ppoint_k+\vect{t})^2}\,,
\end{equation}
where $w_k$ is a weight and $d_\surf{S}(\ppoint)$ denotes the (possibly signed) distance from the point $\ppoint$ to the surface:
\begin{equation}
\left\vert d_\surf{S}(\ppoint)\right\vert = \min_{\ppoint_s\in\surf{S}}{\Vert\ppoint-\ppoint_s\Vert}\,.
\end{equation}
An energy of this form is usually minimized using Gauss-Newton or Levenberg-Marquardt optimization, which need $d_\surf{S}$ and $\nabla d_\surf{S}$ in every iteration of the algorithm.
To emphasize the benefits of Gradient-SDF, we briefly review how the two most common approaches---the iterative closest point (ICP) algorithm, and direct SDF tracking---estimate these quantities.

\paragraph{ICP-based tracking}

If the surface $\surf{S}$ is represented by a point or surfel cloud, \ie, using a discrete set of points $\qpoint_l\in\surf{S}$, possibly with normals $\vect{n}_l$,
the distance $d_\surf{S}$ and its gradient can be estimated either using the (unsigned) \emph{point-to-point} approximation
\begin{align}
    d_\surf{S}^\text{pt-pt}(\ppoint) &= \Vert\ppoint-\qpoint_{l^*}\Vert\,, \\
    \nabla d_\surf{S}^\text{pt-pt}(\ppoint) &= \frac{\ppoint-\qpoint_{l^*}}{\Vert\ppoint-\qpoint_{l^*}\Vert}\,,
\end{align}
or the (signed) \emph{point-to-plane} approximation
\begin{align}
    d_\surf{S}^\text{pt-pl}(\ppoint) &= \vect{n}_{l^*}^\top(\ppoint-\qpoint_{l^*})\,, \\
    \nabla d_\surf{S}^\text{pt-pl}(\ppoint) &= \vect{n}_{l^*}\,,
\end{align}
with
\begin{equation}
    l^* = \arg\min_{l}\Vert\ppoint-\qpoint_l\Vert\,.
\end{equation}
Both require a nearest neighbor search for each evaluation of $d_\surf{S}$ and $\nabla d_\surf{S}$, which can be implemented using a kD-tree, or more efficiently by searching the pixel neighborhood in the depth image.
In approaches like~\cite{newcombe2011kinectfusion,niessner2013real}, where a discrete SDF representation is used to store the global 3D model, each camera pose estimation step needs to convert the SDF representation to a point cloud in order to apply ICP, which is inconsistent and not elegant.

\paragraph{Direct SDF tracking}

Approaches like~\cite{bylow2013real,canelhas2013sdf}, by contrast, directly use the SDF voxels $(\vect{v}_j,\psi_j)$ to approximate $d_\surf{S}$ using interpolation:
\begin{equation}
\label{eq:dir_dist}
    d_\surf{S}^\text{sdf}(\ppoint) = \sum_{\vect{v}_j\in\mathcal{N}(\ppoint)}{\psi_j\omega(\ppoint,\vect{v}_j)}\,,
\end{equation}
where $\mathcal{N}(\ppoint)$ is a neighborhood of $\ppoint$, and $\omega(\ppoint,\vect{v}_j)$ are interpolation coefficients.
For tri-linear interpolation, $\vert\mathcal{N}(\ppoint)\vert = 8$.
The gradient of $d_\surf{S}^\text{sdf}$ can be computed by finite differences over the regular grid of sample points $\vect{v}_j$:
\begin{equation}
\label{eq:dir_grad}
    \nabla d_\surf{S}^\text{sdf}(\ppoint) = \sum_{\vect{v}_j\in\mathcal{N}'(\ppoint)}{\psi_j\begin{pmatrix}\tau_x(\ppoint,\vect{v}_j) \\ \tau_y(\ppoint,\vect{v}_j) \\ \tau_z(\ppoint,\vect{v}_j) \end{pmatrix}}\,.
\end{equation}
The coefficients $\tau_{x,y,z}(\ppoint,\vect{v}_j)$ and the neighborhood $\mathcal{N}'(\ppoint)$ depend on which type of interpolation and which type of finite difference scheme is chosen.

The advantage of such direct approaches over ICP-based ones is that the same volumetric representation that is used for mapping can also be used for pose estimation without any conversion, resulting in a very elegant solution that is easy to implement.
However, to evaluate \eqref{eq:dir_dist} and \eqref{eq:dir_grad} at least 8 voxels need to be read in order to get $d_\surf{S}^\text{sdf}$ and its gradient.
This is most efficient when voxels are stored contiguously in memory, which restricts the reconstruction volume to small to medium sizes (typically at most $512^3$).

\paragraph{Tracking using Gradient-SDF}

With our data structure, we can easily approximate both $d_\surf{S}$  and $\nabla d_\surf{S}$ with only one single voxel look-up, using a first-order Taylor expansion:
\begin{align}
\label{eq:taylor}
    d_\surf{S}^\text{our}(\ppoint) &= \psi_{0} + (\ppoint-\vect{v}_{j^*})^\top\hat{\vect{g}}_{j^*}\,, \\
    \nabla d_\surf{S}^\text{our}(\ppoint) &= \hat{\vect{g}}_{j^*}\,, \\
    j^* = \arg&\min_j \Vert\ppoint-\vect{v}_j\Vert\,.
\end{align}
This looks very similar to the ICP-based formulation, but in our case $j^*$ can be computed without any neighbor search simply by rounding $\ppoint/v_s$, as we know that the $\vect{v}_j$ are sampled on a regular grid in $\mathbb{R}^3$.

As a consequence, contiguous memory storage that is so beneficial for volumetric direct SDF tracking approaches is no longer as important, and we can use a hash map instead to compactly store our voxels, while still staying within one geometry representation.
This allows us to store larger volumes just like in~\cite{niessner2013real}, where voxels far from the surface (\ie with zero weight) are not explicitly stored.

\subsection{Pose optimization and bundle adjustment}

Typically, bundle adjustment is performed on a sparse set of points~\cite{triggs1999bundle, agarwal2010bundle}, as computational cost grows with the number of points.
In online approaches, also the number of cameras is usually limited to a sliding window.
With the introduction of BAD SLAM~\cite{schops2019bad}, these limitations are lifted:
the use of depth data for bundle adjustment together with some smart optimization allows for actually performing bundle adjustment on a more dense level.
Naturally the question comes up if bundle adjustment can also be performed in implicit dense geometry representations such as signed distance fields.
This is where Gradient-SDF comes in handy:
while it is very hard to come up with a meaningful bundle adjustment energy in standard SDF representations, we can exploit \eqref{eq:grad_sdf_project} to define points on the surface for which we want to adjust bundles.
Together with the finding of~\cite{schops2019bad} that optimization can be limited to the normal direction, we can set up an \emph{implicit photometric BA} cost:
\begin{equation}
\label{eq:rgb_cost}
    E(\{\mat{R}_i, \vect{t}_i\}, \psi) =
    \sum_{i,j,c}{\nu_{ij}\Phi\bigl(I_{ij}^c- \tfrac{1}{N_j}\sum_i{\nu_{ij}I_{ij}^c}\bigr)}\,,
\end{equation}
where $\nu_{ij}$ denotes the visibility of voxel $\vect{v}_j$ in frame $i$ ($N_j=\sum_i{\nu_{ij}}$), $c\in\{\text{r},\text{g},\text{b}\}$, and $\Phi$ is a robust cost function.
$I_{ij}^c$ is given by
\begin{equation}
\label{eq:intensity_proj}
    I_{ij}^c(\{\mat{R}_i, \vect{t}_i\}, \psi_j) =
    I_i^c\left( \pi(\mat{R}_i^\top(\vect{v}_j-\psi_j\hat{\vect{g}}_j-\vect{t}_i))\right)\:,
\end{equation}
with $\pi$ the perspective projection from $\mathbb{R}^3$ to the image domain.
In the optimization, we abstract the original meaning of $\hat{\vect{g}}_j$ as gradient of $\psi_j$, and keep it fixed while changing $\psi_j$.
Following~\cite{schops2019bad}, for strongly connected problems, the optimization of poses and distances can be performed alternatingly to reduce computational cost.
We can limit the pose optimization part to voxels that actually contain surface points to reduce computations.
Depending on the scene, this approach can be further accelerated by limiting the optimization to the camera poses $(\mat{R}_i,\vect{t}_i)$.
Rather than projecting voxel centers $\vect{v}_j$ into the RGB images $I_i$ (as done in \eg \cite{bylow2013real,niessner2013real}), we project the real surface point $\vect{v}_j-\psi_j\hat{\vect{g}}_j$.
This is not easily possible in a standard SDF representation, and while the effect for simple BA optimization is small, it opens up new applications such as high-resolution surface optimization using shading and lighting information to improve approaches like~\cite{bylow2019combining}.
Gradient-SDF thus allows for (photometric) bundle adjustment in an implicit voxel-based representation.

Different cost formulations exist for photometric BA, and we chose \eqref{eq:rgb_cost} which is similar to~\cite{delaunoy2014photometric} rather than a sum of pairwise squared intensity differences, because we appreciate the natural interpretation that this provides:
for $\Phi(r)=r^2$, the energy $E$ can be rewritten as
\begin{equation}
    \label{eq:variance}
    E(\{\mat{R}_i, \vect{t}_i\}, \psi) = \sum_{j,c}{N_j\cdot \operatorname{Var}(\{I_{ij}^c\})}\,,
\end{equation}
a weighted sum of the intensity variance of each voxel's closest surface point.
As we do not store the color per voxel in contrast to \cite{schops2019bad}, \eqref{eq:rgb_cost} couples all camera poses, making pose optimization quadratically dependent on the number of frames used.
We lift this limitation by decoupling the original energy cost and simultaneously minimizing
\begin{equation}
\label{eq:rgb_cost_pose}
    E_i(\mat{R}_i, \vect{t}_i, \psi) =
    \sum_{j,c}{\nu_{ij}\Phi\bigl(I_{ij}^c- \tfrac{1}{N_j}\sum_i{\nu_{ij}I_{ij}^c}\bigr)}\,.
\end{equation}
This makes the pose optimization step linear in the number of frames.
The simultaneous (rather than alternating) minimization of pose energies means that even though each of the energies contains an average of all residuals $I_{ij}^c$, this average does not impose any additional computation in practice.
We show in the evaluation that the simplification introduced in \eqref{eq:rgb_cost_pose} has nearly no effect on results.
After minimizing $E$ w.r.t. $\psi_j$ and $(\mat{R}_i,\vect{t}_i)$, we can compute the color $\rho_j^c$ at voxel $\vect{v}_j$ as the mean
\begin{equation}
\label{eq:mean_color}
    \rho_j^c = \frac{1}{N_j}\sum_{i}{\nu_{ij}I_{ij}^c}%
\end{equation}

Since we have the projection to the closest surface point implicitly encoded in the cost function \eqref{eq:rgb_cost}, and do not store an explicit estimate of voxel color, we avoid the correspondence estimation between geometry and texture that is used in TextureFusion~\cite{lee2020texturefusion}.

\subsection{Surface extraction from a Gradient-SDF}
\label{sec:surface_extraction}

In order to eventually extract a surface from the implicit TSDF representation, we have two choices:
we can extract a set of surfels, or we run marching cubes to extract a mesh.
We discuss both approaches and present ways to efficiently implement them given our specific data structure.

\paragraph{Oriented point cloud extraction}

A very fast way to extract a surface representation from our gradient-augmented SDF representation is to extract surface points $\ppoint_s(\vect{v}_j)$ together with their normals $-\hat{\vect{g}}_j$ from all voxels that have $|\psi_j\hat{\vect{g}}_j|\leq\frac{v_s}{2}$ (component-wise).
This results in a homogeneously sampled, consistently oriented point cloud of resolution $\frac{v_s}{2}$.
Re-sampling for more regularly distributed surfels such as in~\cite{whelan2015real} is not necessary in our Gradient-SDF representation.
Regular geometry upsampling of this point cloud is also very easy:
we subdivide each voxel into four subvoxels, determine their distance using the Taylor expansion \eqref{eq:taylor}, and then extract surfels from subvoxels with $|\psi_j\hat{\vect{g}}_j|\leq\frac{v_s}{4}$ instead.

\paragraph{Layered Marching Cubes for mesh extraction}

Extracting a mesh rather than a set of surfels is also easy for an implicit SDF representation like ours:
we can use the well-known marching cubes (MC) algorithm~\cite{lorensen1987marching}.
In the case where voxels are not stored volumetrically in memory, but in a hash map, we can traverse the hash map and check for each voxel if the eight corners of the cube next to it (going in positive $x$-, $y$- and $z$-direction) are allocated.
If yes, we can extract the corresponding triangle face from the MC lookup table.
This, however, requires a complete re-implementation of the meshing algorithm.
We use a different approach that can be more seamlessly integrated in an existing MC implementation:
in our \emph{layered marching cubes}, we first extract the minimum and maximum $x$, $y$ and $z$ coordinates of all voxels to obtain an axis-aligned bounding box.
Then, starting from (without loss of generality) minimal $z$, we allocate memory for two voxel layers of size $(x_{\max}-x_{\min})\times(y_{\max}-y_{\min})$, and fill it with the first two $z$-layers for weights, colors, and distance values.
We now apply Marching Cubes on the layer interface.
After this, we re-fill the first layer with the next $z$-values and proceed.
This way, we avoid a cubic memory usage.

\section{Evaluation}

\begin{figure}
    \centering
    \includegraphics[height=.48\linewidth, angle=-90]{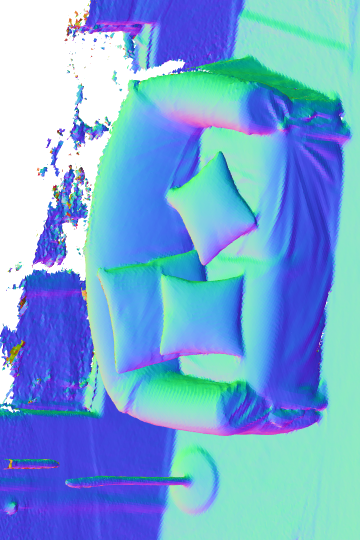}
    \hfill
    \includegraphics[height=.48\linewidth, angle=-90]{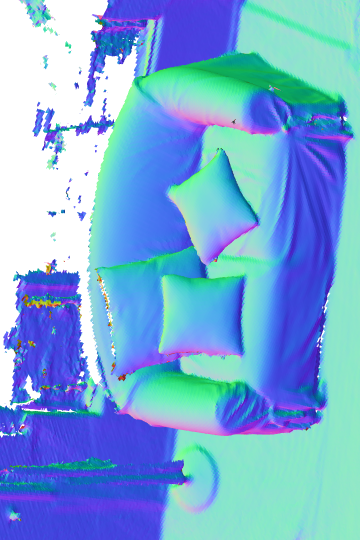}
    \vspace{1.5mm}
    \caption{Qualitative reconstruction results for SDF Tracker~\cite{canelhas2013sdf} with $512^3=1.34\times 10^8$ voxels (\emph{left}) and our tracker using a Gradient-SDF hash map (\emph{right}) on the \emph{00577\_sofa} sequence from~\cite{choi2016large} .
    Shown are the meshes after running marching cubes.
    The visual quality of results is largely comparable, but our hash map-based method needs more than $20\times$ less memory than the dense storage of~\cite{canelhas2013sdf}, despite having three more entries per voxel.
    }
    \label{fig:tracking_sofa}
\end{figure}

To demonstrate the potential of Gradient-SDF, we analyse our stored gradients and show two example applications where we use Gradient-SDF to store the underlying 3D geometry:
a simple tracking and mapping system using depth images, and photometric bundle adjustment together with subsampling on a Gradient-SDF initialized from our tracking system.

\subsection{Implementation}

We will make our code available open-sourced after publication.
Our data structure is implemented in C++ with single precision \texttt{float}s, and our \texttt{GradSdfVoxel}s are stored in a hash map, one voxel per entry.
We found that for a CPU implementation, the difference in performance compared to hashing blocks of size $8^3$ is marginal.
For pose optimization both in the tracker and in the bundle adjustment, we use Gauss-Newton and solve the resulting linear system using Cholesky decomposition.
Optimizing the BA cost \eqref{eq:rgb_cost_pose}, we do not store the three variables for voxel color, which is in contrast to~\cite{bylow2013real, niessner2013real}.
All experiments were performed on an Intel Xeon CPU @ $3.60\,\text{GHz}$, using OpenMP with four threads, and no GPU.

\subsection{Gradient quality on synthetic data}

\paragraph{Setup}
We start our evaluation with an analysis of the gradients $\hat{\vect{g}}_j$ that we store in our voxels, and compare them and finite difference gradients to ground truth gradients.
Five random spheres with different radii are randomly rendered into a sequence of depth images augmented with Kinect-like sensor noise~\cite{khoshelham2012accuracy}.
We perform (Gradient-)SDF fusion according to the formulas in \eqref{eq:sdf_update}--\eqref{eq:gradient_update} using ground truth poses.
For a sphere with center $\vect{c}$, the (ground truth) SDF gradient at point $\ppoint$ is $\frac{\ppoint-\vect{c}}{\Vert\ppoint-\vect{c}\Vert}$.
Furthermore, we have the stored gradient and we can compute a central finite difference gradient from the accumulated SDF.

\begin{figure}
    \centering
    \includegraphics[width=\linewidth]{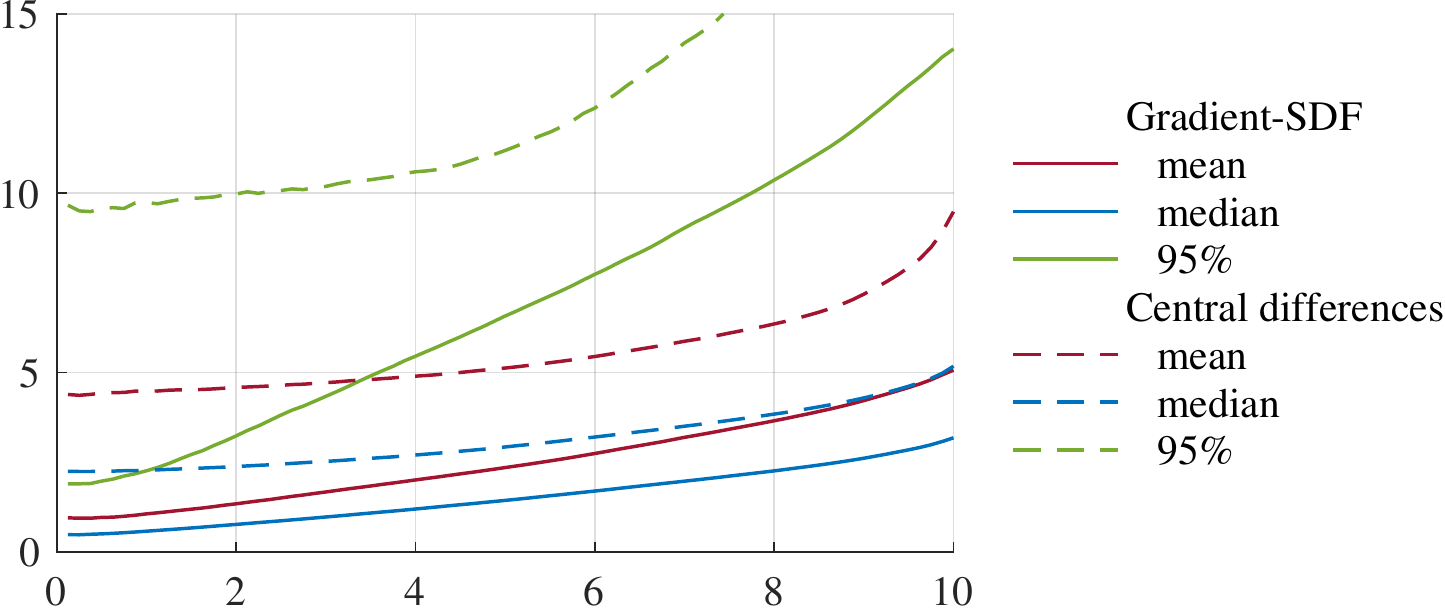}
    \caption{Quality of gradient estimates.
    For all voxels closer than $x$ voxels to the surface, the $y$-value of the curves specify mean, median and $95^\text{th}$ percentile of the angular deviation from ground truth gradients in degrees.
    \emph{Solid lines} are Gradient-SDF vectors, and \emph{dashed lines} central finite differences.
    Our gradients are significantly more accurate than those computed using finite differences, \eg the mean angular deviation of voxels within $10v_s$ from the surface is nearly twice as big for central differences ($9.49^\circ$) than for our stored gradients ($5.07^\circ$).
    }
    \label{fig:gradient_analysis}
\end{figure}

\paragraph{Results}
Figure~\ref{fig:gradient_analysis} shows the angular deviation of gradient vectors to ground truth, for our stored gradients and for finite differences.
Our gradients are clearly much closer to the ground truth ones---both for surface voxels, and throughout the whole truncation region.
Additional visualizations in the Appendix show that our gradient estimate is much smoother than the one using finite differences.
Thus, not only does Gradient-SDF enable easier interpolation of $d_\surf{S}(\ppoint)$, it also produces much better gradient estimates than a standard finite difference scheme.

\subsection{Camera tracking using depth images}

\begin{table}[t]
    \centering
    \begin{tabular}{l r r r r r r}
         & \rotatebox{90}{KinFu~\cite{newcombe2011kinectfusion}} 
         & \rotatebox{90}{SDF-Fusion~\cite{bylow2013real}}
         & \rotatebox{90}{\parbox{1.9cm}{SDF tracking hash map}}
         & \rotatebox{90}{\parbox{2.25cm}{Gradient-SDF hash map (ours)}} \\
    \midrule
        fr1/desk & 6.8 & 3.5 & 3.9 & 5.6 \\
        fr1/desk2 & \textcolor{mlred}{63.5} & 6.2 & 6.6 & 6.6 \\
        fr1/xyz & 2.5 & 2.3 & 1.7 & 2.0 \\
        fr1/rpy & 8.1 & 4.2 & 3.9 & 4.9 \\
        fr1/plant & \textcolor{mlred}{28.1} & 4.3 & 5.5 & 11.2 \\
        fr1/teddy & \textcolor{mlred}{33.7} & 8.0 & 10.1 & 11.3 \\
        fr3/household & 6.1 & 4.0 & 3.8 & 5.2 \\
    \bottomrule
    \end{tabular}
    \vspace{1.5mm}
    \caption{RMSE of the absolute trajectory error (ATE) in cm on sequences from~\cite{sturm2012benchmark}, errors above $25\,\text{cm}$ are marked \textcolor{mlred}{red}.
    On \emph{fr1/floor}, all methods have an ATE above $50\,\text{cm}$, thus we exluded that sequence.
    While very slow, in terms of ATE our baseline implementation of direct SDF tracking using a hash map is on par with~\cite{bylow2013real}.
    Gradient-SDF is much faster and still consistently outperforms KinectFusion and is comparable to standard direct SDF tracking.
    Results for KinFu and the direct tracker are taken from~\cite{bylow2013real}.
    }
    \label{tab:tracking_results}
\end{table}

Our first real-world application is a 3D scanning system that uses depth images to track the camera and build up an implicit 3D model, just like KinectFusion or direct SDF tracking.

\paragraph{Setup}
We take the general system setup from~\cite{bylow2013real}, with a linear weight, and cut off depth values at $3.5\text{m}$.
We choose a voxel size of $v_s=2\,\text{cm}$ and truncated at $5v_s$.
Since a hash map representation does not allow for a straightforward voxel-wise SDF update~\cite{niessner2013real}, we update voxels based on depth image pixels:
for each pixel, we update all voxels along the viewing ray that are within the truncation distance.
In case the pixel size at the given depth is larger than one voxel, this can lead to some voxels not being updated, thus it is important to not choose the voxel size too small.
Normals are estimated using the FALS method from~\cite{badino2011fast}, and only points whose angle between normal and viewing ray is less than $75^\circ$ are integrated into the final Gradient-SDF.
To have a fair comparison against KinectFusion and direct SDF tracking, we do not use the color values as in~\cite{niessner2013real}.
In order to evaluate the benefits compared to a direct SDF tracker with a hash map implementation, we implemented such a tracker as a baseline.
This hash map SDF tracker stores voxels sparsely, but still uses tri-linear interpolation to evaluate $d_\surf{S}$ and $\nabla d_\surf{S}$, which means each function/gradient evaluation needs eight hash table lookups.

\paragraph{Results}
In Table~\ref{tab:tracking_results}, we summarize results on the TUM RGB-D dataset~\cite{sturm2012benchmark}.
To make this quantitative evaluation reproducible, we switched off OpenMP to generate the table.
\cite{bylow2013real,newcombe2011kinectfusion} on average perform best for a $512^3$ voxel reconstruction volume~\cite{bylow2013real}, so we report numbers for that setting.
In terms of average pose error, using a Gradient-SDF and interpolating using Taylor expansion rather than tri-linear interpolation is comparable to the other methods, while being superior in terms of representation consistency for sparse storage schemes.
In addition, we show qualitative results in Figure~\ref{fig:tracking_sofa} and the Appendix.
For this qualitative comparison, we use the open-sourced SDF tracker code with a volume of $512^3$ voxels, and the same parameters as our Gradient-SDF tracker.
The SDF tracker implementations has been extended to give access to the SDF volume, which allows us to run marching cubes on the raw data.

\paragraph{Runtime and memory}
Pose estimation takes about $30$--$40\,\text{ms}$ per frame, compared to $100$--$120\,\text{ms}$ for the hash map direct SDF tracker we implemented as baseline.
This clearly shows the superiority of Taylor interpolation over tri-linear for sparse storage schemes.
Integration into the SDF volume takes on average $80\,\text{ms}$ and is about $7\%$ ($5\,\text{ms}$) slower than the baseline implementation which does not integrate the gradients.
Thus, the gains in tracking more than compensate for the marginal overhead introduced in the mapping phase, and a GPU implementation of Gradient-SDF has the potential to run in real time.

The dense reconstruction volume of \cite{bylow2013real} and \cite{newcombe2011kinectfusion} consists of $512^3$ voxels, which means memory for $1.34\times 10^8$ voxels and thus $2.68\times 10^8$ floating point variables (weights and distances per voxel) is required.
This is in contrast to on average $2\times 10^6$ voxels for hash map-based storage.
Our baseline hash map direct SDF tracker thus needs about $4\times 10^6$ floats.
Gradient-SDF stores gradients in addition, so it has in total five values per voxel, resulting in about $10^7$ variables.
Despite one single voxel being $2.5\times$ larger than for a standard SDF, in total this is still more than $20\times$ less memory than volumetric storage.
Additionally, for volumetric storage we need to know in advance where the reconstruction volume shall be placed, making those approaches less flexible in unknown environments.

\subsection{Bundle adjustment and pose optimization}

\begin{figure}
    \newcommand{\imgwidth}{0.75\linewidth}
    \newcommand{\imagewithtitle}[2]{
        \begin{tikzpicture}[baseline=0]
        \begin{scope}[local bounding box=img]
            \clip (-0.5*\imgwidth, -0.25*\imgwidth) rectangle (0.5*\imgwidth, 0.25*\imgwidth);
            \node {\includegraphics[width=\imgwidth, angle=180]{#1}};
        \end{scope}
            \node[anchor=north, align=left] at ($(img.north) + (0, 0.25)$) {\footnotesize{#2}};
        \end{tikzpicture}
    }
    
    \centering
    
    \imagewithtitle{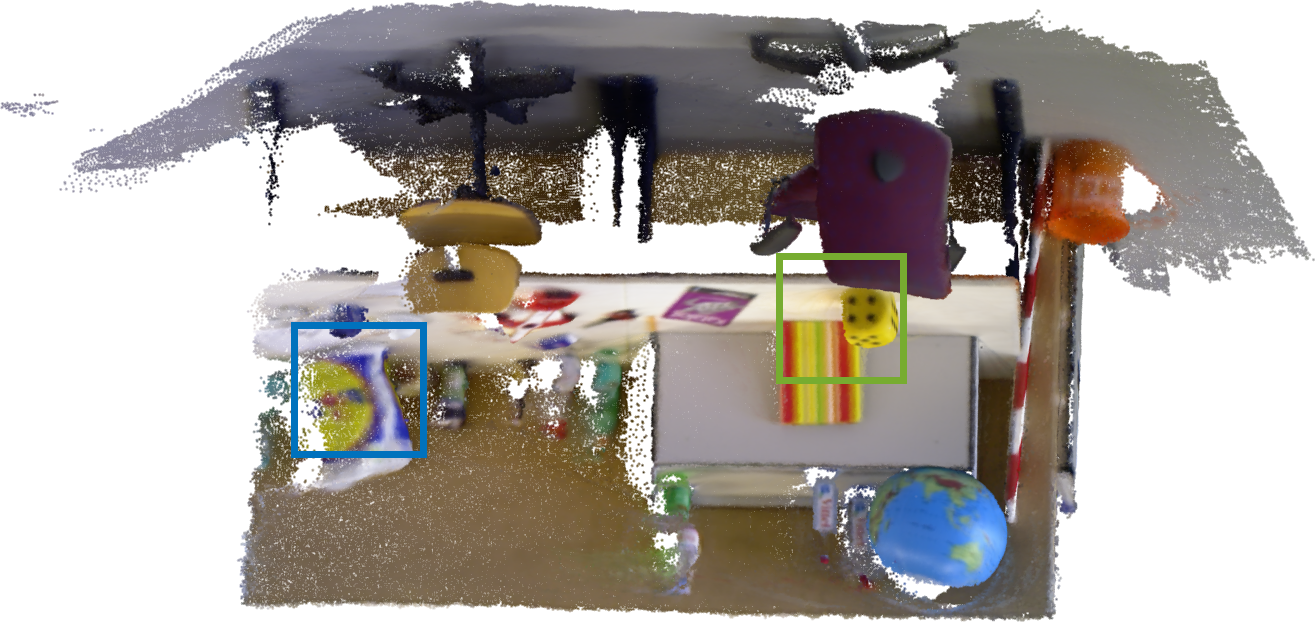}{BAD SLAM, (630K points, RMSE 1.5\,cm)}
    \begin{minipage}{0.2\linewidth}
    \centering
    \includegraphics[width=0.9\linewidth,cframe=mlgreen 1.5pt 0pt, angle=180]{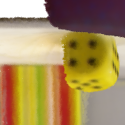}
    \includegraphics[width=0.9\linewidth,cframe=mlblue 1.5pt 0pt, angle=180]{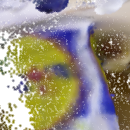}
    \end{minipage}
    \imagewithtitle{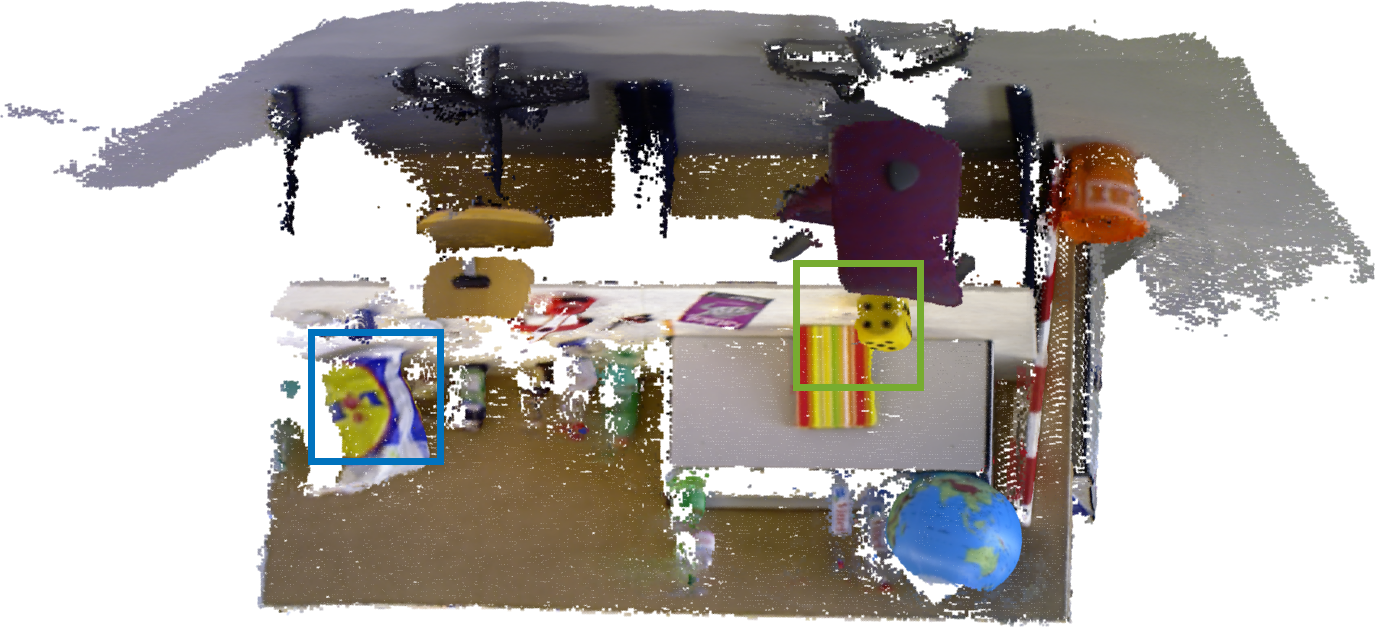}{Gradient-SDF final (470K points, RMSE 0.6\,cm)}
    \begin{minipage}{0.2\linewidth}
    \centering
    \includegraphics[width=0.9\linewidth,cframe=mlgreen 1.5pt 0pt, angle=180]{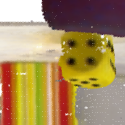}
    \includegraphics[width=0.9\linewidth,cframe=mlblue 1.5pt 0pt, angle=180]{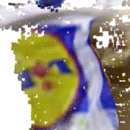}
    \end{minipage}

    \vspace{1.5mm}
    \caption{Colored point cloud produced by BAD SLAM (\emph{top}), and %
    after optimization of the BA cost on our Gradient-SDF (\emph{bottom}), \emph{fr3/long\_office\_household}~\cite{sturm2012benchmark}: the very low root mean squared pose error becomes apparent on the reprojected average colors of the surface points, for our Gradient-SDF even more than for BAD SLAM.}
    \label{fig:ba_results_household}
\end{figure}
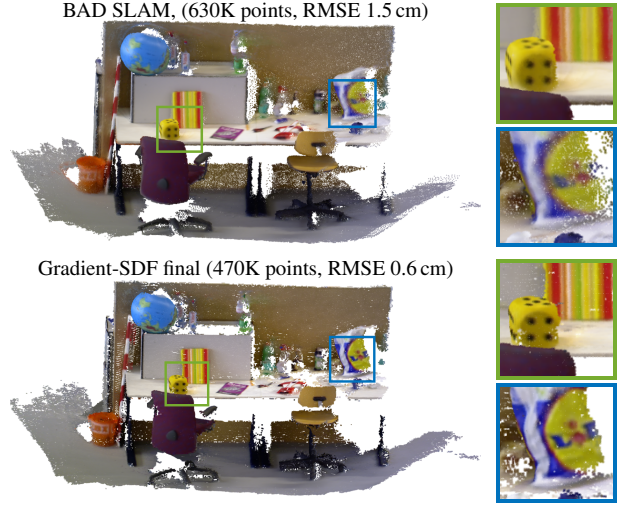

Our second set of experiments demonstrates the power of our implicit photometric bundle adjustment/pose optimization formulation:
on different sequences from~\cite{sturm2012benchmark}, we minimize the bundle adjustment cost from \eqref{eq:rgb_cost} for 30 keyframes over 10 seconds, \ie a keyframe ratio of $10\%$, as in BAD SLAM, and a regularizer weight of $0.01\text{cm}^{-2}$.
The initial geometry is obtained from our depth-based tracker.

\begin{table}[t]
    \centering
    \begin{tabular}{l r r r r r }
        & \rotatebox{90}{fr1/xyz}
        & \rotatebox{90}{fr1/teddy}
        & \rotatebox{90}{fr2/xyz}
        & \rotatebox{90}{fr2/rpy}
        & \rotatebox{90}{fr3/household} \\
      \midrule
      before optimization & 2.8 & 6.8 & 1.8 & 2.7 & 2.1 \\
      \midrule
        full BA, poses coupled & 1.0 & 4.4 & 1.0 & 2.5 & 0.6 \\
        pose only, poses coupled & 1.0 & 4.3 & 1.1 & 2.4 & 0.7 \\
        full BA, decoupled & 1.2 & 4.6 & 1.0 & 2.5 & 0.6 \\
        pose only, decoupled & 1.1 & 4.3 & 1.0 & 2.4 & 0.6 \\
      \midrule
        BAD SLAM~\cite{schops2019bad} & 1.8 & 3.7 & 1.3 & 0.9 & 1.5 \\
      \bottomrule
    \end{tabular}
    \vspace{1.5mm}
    \caption{RMSE [cm] of pose error translation on 30 keyframes, sequences from~\cite{sturm2012benchmark}:
    even with a number of computational optimizations (see main text), we are able to keep the error in our implicit bundle adjustment very low compared to the initial error after depth tracking.
    For reference, we provide numbers for BAD SLAM as an example of an explicit dense bundle adjustment approach.
    Overall, our errors are comparable to those of BAD SLAM.}
    \label{tab:pose_opt}
\end{table}

\paragraph{Quantitative results}
All results are summarized in Table~\ref{tab:pose_opt}:
We first perform photometric BA by minimizing \eqref{eq:rgb_cost}.
Poses improve substantially after optimization, while geometry only changes marginally, which we attribute to the good initial estimate that our depth tracker provides.
Thus, we run a version with pose-only optimization, which is faster and achieves on-par pose errors, plus looks equally good visually (note that we do not have ground truth for geometry).
We perform the same set of experiments for pose optimization decoupled as in \eqref{eq:rgb_cost_pose}, and find that we get nearly the same improvement on poses, while keeping the computational complexity linear in the number of keyframes.
With these optimizations, in total we can minimize the BA cost in $20$--$30\,\text{ms}$ per iteration and pose using a single-threaded CPU implementation, meaning there is good potential for real-time capability on a GPU.
As surfel-based reference method, we run BAD SLAM and use their pose estimates for quantitative evaluation.

\begin{figure}
    \raggedleft
    
    \newcommand{\imgwidth}{0.65\linewidth}
    \newcommand{\imagewithzoomandtitle}[4]{
        \begin{tikzpicture}[baseline=0]
        \begin{scope}[local bounding box=img]
            \clip (-0.5*\imgwidth, -0.45*\imgwidth) rectangle (0.5*\imgwidth, 0.45*\imgwidth);
            \node {\includegraphics[width=0.55\linewidth, angle=60]{#1}};
        \end{scope}
        \node[anchor=west, inner sep=0pt, outer sep=0pt] (zoom) at ($(img.east) + (0.6, 0.0)$) {%
            \begin{minipage}{0.2\linewidth}
                \includegraphics[width=0.9\linewidth,cframe=mlgreen 1.5pt 0pt, angle=90]{#2}
                \includegraphics[width=0.9\linewidth,cframe=mlblue 1.5pt 0pt, angle=90]{#3}
            \end{minipage}};
        \node[anchor=south east, align=right, font=\footnotesize] at ($(zoom.north east) + (0, 0.0)$) {#4};
        \end{tikzpicture}
    }
    
    \imagewithzoomandtitle{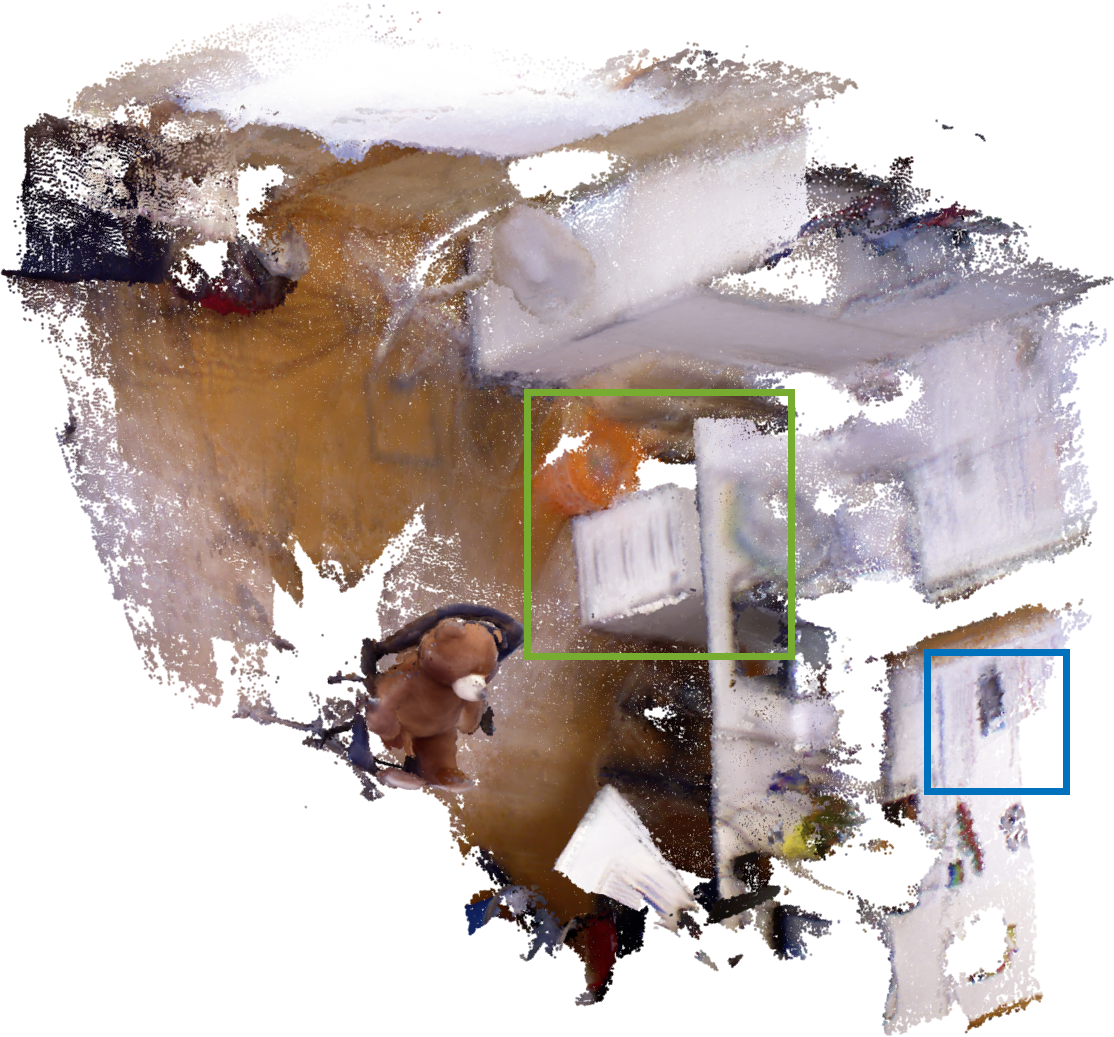}{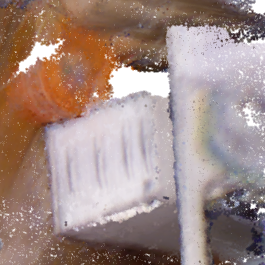}{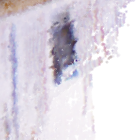}{BAD SLAM\\(800K points, RMSE 3.7\,cm)}

    \vspace{-0.3cm}
    \imagewithzoomandtitle{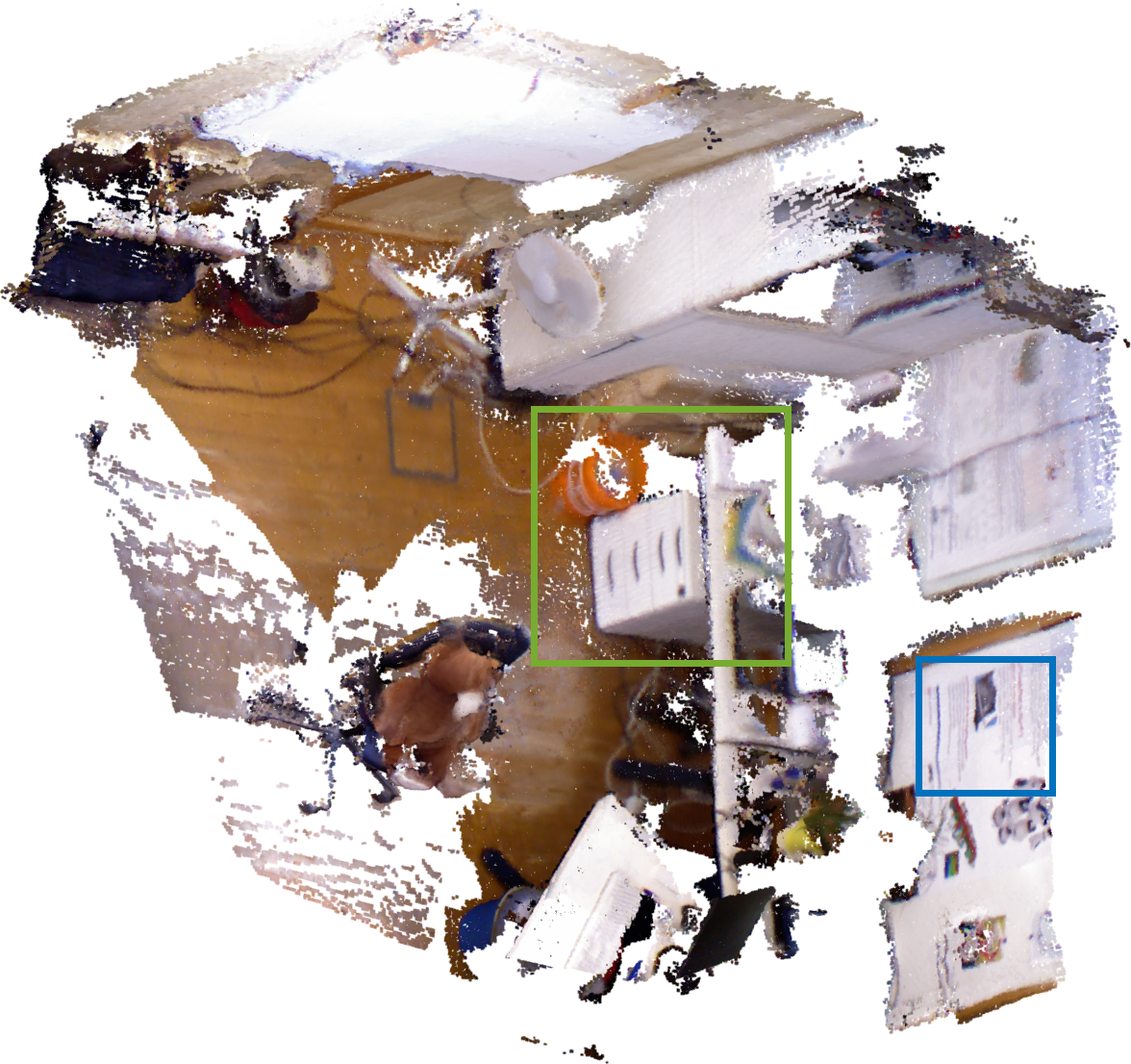}{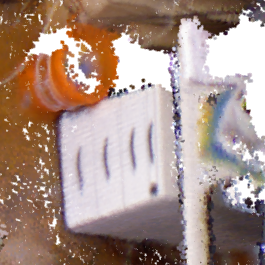}{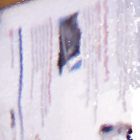}{Gradient-SDF final\\(1M points, RMSE 4.3\,cm)}
    
    \vspace{-0.3cm}
    \caption{Colored point cloud produced by BAD SLAM (\emph{top}), and %
    after optimization of the BA cost on our Gradient-SDF (\emph{bottom}), \emph{fr1/teddy}~\cite{sturm2012benchmark}: despite slightly larger root mean squared pose error, our final texture is much sharper than that of BAD SLAM.}
    \label{fig:ba_results_teddy}
\end{figure}
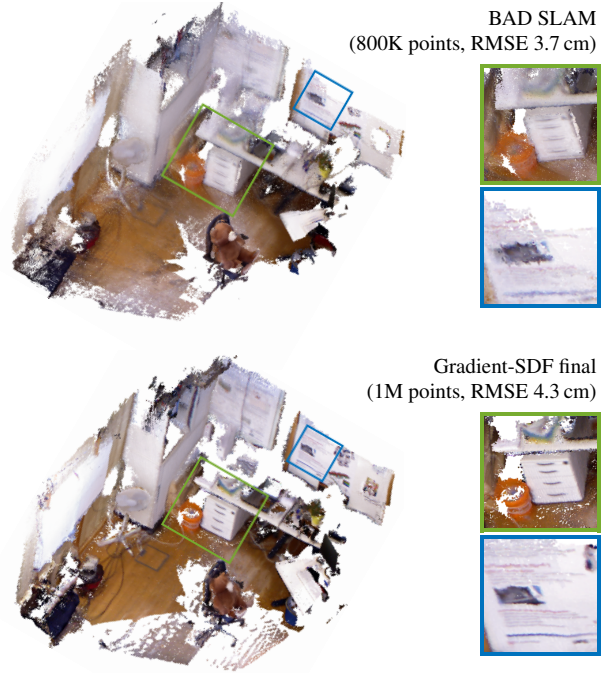

\paragraph{Qualitative results}
We show qualitative results for our photometric optimization (using decoupled poses) and BAD SLAM's built-in surface reconstruction, again with the 30 keyframes out of 300 frames, in Figures \ref{fig:ba_results_household} (\emph{fr3/household}) and \ref{fig:ba_results_teddy} (\emph{fr1/teddy}).
We extract the geometry in double resolution as outlined in \ref{sec:surface_extraction} to get a denser point cloud that has a number of points comparable to the BAD SLAM result.
For other sequences, see Appendix.
The better poses and distances are adjusted, the lower the variance of the mean reprojected color in \eqref{eq:mean_color} will be, and thus the sharper the texture.
For this reason, we show colored point clouds for qualitative visualization.
Since we have depth and color input decoupled -- one is used for the tracker only, one for the photometric optimization, we nowhere assume synchronized data and can obtain very sharp textures even for unsynchronized datasets like TUM RGB-D~\cite{sturm2012benchmark}.
This is in contrast to BAD SLAM and many other RGB-D scanning systems, which assume synchronized data in their model.

\section{Discussion}

While the memory overhead compared to hash map-based SDFs without gradients is not critical for a typical indoor scene, we still aim to reduce memory.
Our next step is thus to exploit the fact that unit norm gradients are in $\mathbb{S}^2$.
We will investigate how to parameterize $\hat{\vect{g}}_j$ with two variables and still keep the update rule simple.
The shown applications of Gradient-SDF are exemplary, the representation can be used to store any 3D geometry.
Thus, we kept our set of experiments small and instead focused on the theoretical description of how Gradient-SDF can be used.
In the future we will explore the benefits of our Gradient-SDF for tracking and mapping in neural geometry representations.

\paragraph{Conclusion}
We proposed Gradient-SDF as a hybrid representation for 3D geometry that combines the advantages of explicit and implicit representations.
By enhancing the classical implicit signed distance function (SDF) with the gradient value, we achieve capacities of explicit representations including direct photometric bundle adjustment.
In several experiments we demonstrate these advantages---in particular, our gradients are much more accurate than finite difference approximations, and 3D scanning using Gradient-SDF produces impressively sharp reconstructions. 

{\small
\bibliographystyle{ieee_fullname}
\bibliography{egbib}
}

\onecolumn
\appendix


\renewcommand{\thefigure}{A.\arabic{figure}}
\renewcommand{\thetable}{A.\arabic{table}}
\setcounter{figure}{0}
\setcounter{table}{0}

\section{Gradient quality visualized}

\begin{figure}[H]
\centering
    \makebox[0.19\linewidth][c]{\small{Ground truth}}
    \hfill
    \makebox[0.19\linewidth][c]{\small{Forward differences}}
    \hfill
    \makebox[0.19\linewidth][c]{\small{Backward differences}}
    \hfill
    \makebox[0.19\linewidth][c]{\small{Central differences}}
    \hfill
    \makebox[0.19\linewidth][c]{\small{Gradient-SDF (ours)}}
    \\ \vspace{0.15cm}
\includegraphics[width=0.19\linewidth, frame]{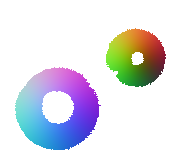}
\hfill
\includegraphics[width=0.19\linewidth, frame]{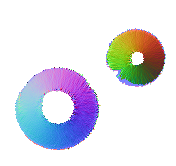}
\hfill
\includegraphics[width=0.19\linewidth, frame]{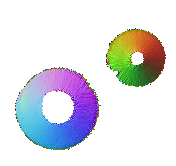}
\hfill
\includegraphics[width=0.19\linewidth, frame]{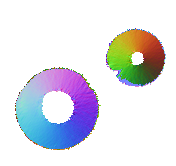}
\hfill
\includegraphics[width=0.19\linewidth, frame]{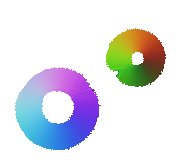}
\\ \vspace{0.15cm}
\includegraphics[width=0.19\linewidth, frame]{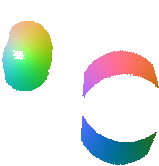}
\hfill
\includegraphics[width=0.19\linewidth, frame]{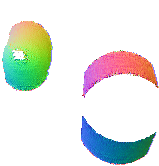}
\hfill
\includegraphics[width=0.19\linewidth, frame]{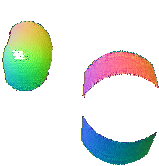}
\hfill
\includegraphics[width=0.19\linewidth, frame]{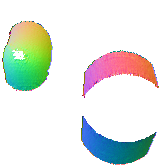}
\hfill
\includegraphics[width=0.19\linewidth, frame]{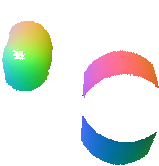}
\\ \vspace{0.15cm}
\includegraphics[width=0.19\linewidth, frame]{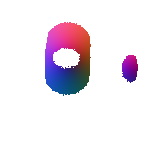}
\hfill
\includegraphics[width=0.19\linewidth, frame]{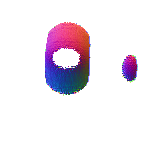}
\hfill
\includegraphics[width=0.19\linewidth, frame]{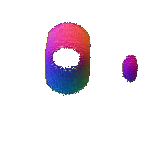}
\hfill
\includegraphics[width=0.19\linewidth, frame]{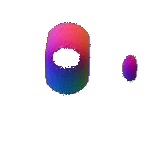}
\hfill
\includegraphics[width=0.19\linewidth, frame]{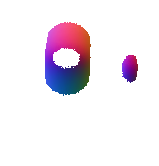}
\\ \vspace{0.15cm}
\includegraphics[width=0.19\linewidth, frame]{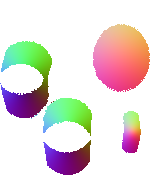}
\hfill
\includegraphics[width=0.19\linewidth, frame]{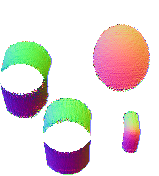}
\hfill
\includegraphics[width=0.19\linewidth, frame]{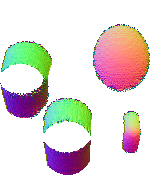}
\hfill
\includegraphics[width=0.19\linewidth, frame]{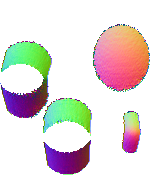}
\hfill
\includegraphics[width=0.19\linewidth, frame]{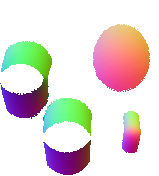}
\\ \vspace{0.15cm}
\includegraphics[width=0.19\linewidth, frame]{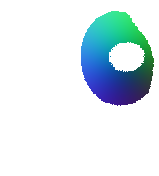}
\hfill
\includegraphics[width=0.19\linewidth, frame]{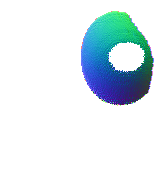}
\hfill
\includegraphics[width=0.19\linewidth, frame]{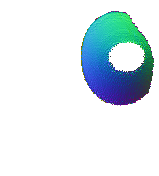}
\hfill
\includegraphics[width=0.19\linewidth, frame]{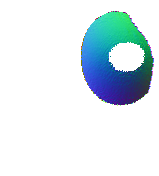}
\hfill
\includegraphics[width=0.19\linewidth, frame]{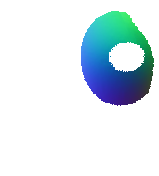}
\\
\caption{Qualitative results for the gradient of the signed distance function of the rendered spheres from Figure~\ref{fig:gradient_analysis}. Shown are five slices through the 3D gradient color-coded as normal map (after normalization) in regions where the computed SDF is non-truncated and has a non-zero weight (truncation is chosen as $10v_s$ for a voxel size $v_s$ of $1\,\text{cm}$).
While central finite differences (\emph{column 4}) are clearly better and less noisy than simple forward or backward differences (\emph{columns 2, 3}), our Gradient-SDF vectors (\emph{right}) are still significantly smoother and less noisy. Comparing to the ground truth (\emph{left}), we can visually confirm the quantitative result from Figure~\ref{fig:gradient_analysis}: our gradients are a lot closer to the ground truth ones than central finite differences.}
\label{fig:sphere_gradients}
\end{figure}

\begin{figure}[H]
\centering
    \makebox[0.3\linewidth][c]{\small{3D view with $z$-slice}}
    \hfill
    \makebox[0.21\linewidth][c]{\small{SDF slice}}
    \hfill
    \makebox[0.21\linewidth][c]{\small{Central differences slice}}
    \hfill
    \makebox[0.21\linewidth][c]{\small{Gradient-SDF slice}}
    \\ \vspace{0.15cm}
\includegraphics[width=0.3\linewidth]{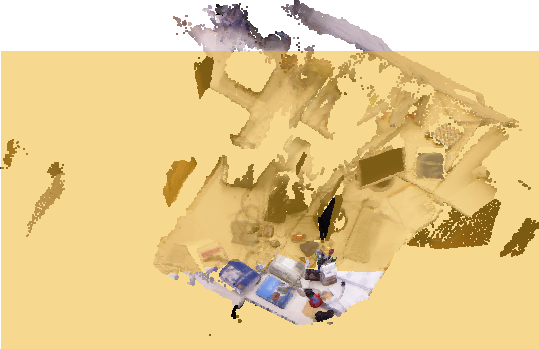}
\hfill
\includegraphics[width=0.22\linewidth, frame]{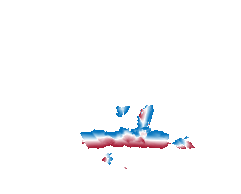}
\hfill
\includegraphics[width=0.22\linewidth, frame]{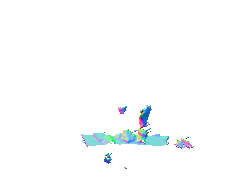}
\hfill
\includegraphics[width=0.22\linewidth, frame]{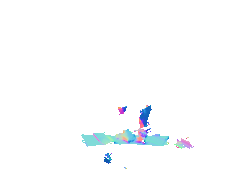}
    \\ \vspace{0.05cm}
\includegraphics[width=0.3\linewidth]{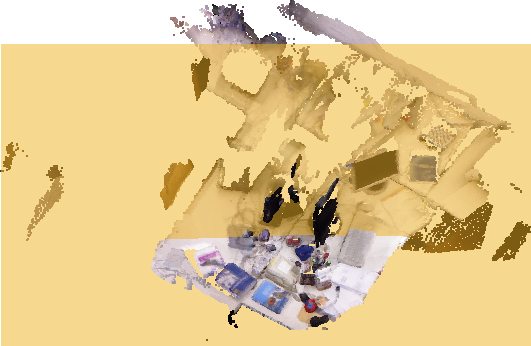}
\hfill
\includegraphics[width=0.22\linewidth, frame]{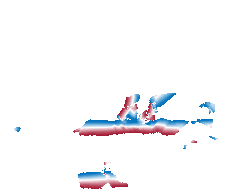}
\hfill
\includegraphics[width=0.22\linewidth, frame]{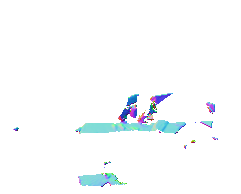}
\hfill
\includegraphics[width=0.22\linewidth, frame]{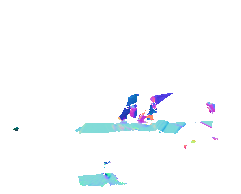}
    \\ \vspace{0.05cm}
\includegraphics[width=0.3\linewidth]{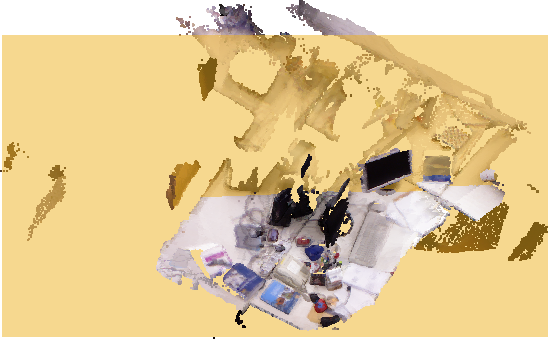}
\hfill
\includegraphics[width=0.22\linewidth, frame]{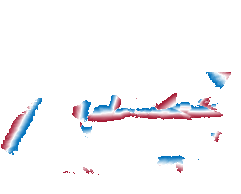}
\hfill
\includegraphics[width=0.22\linewidth, frame]{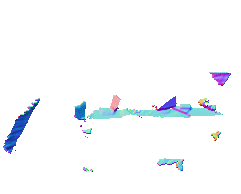}
\hfill
\includegraphics[width=0.22\linewidth, frame]{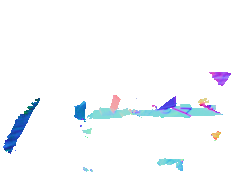}
    \\ \vspace{0.05cm}
\includegraphics[width=0.3\linewidth]{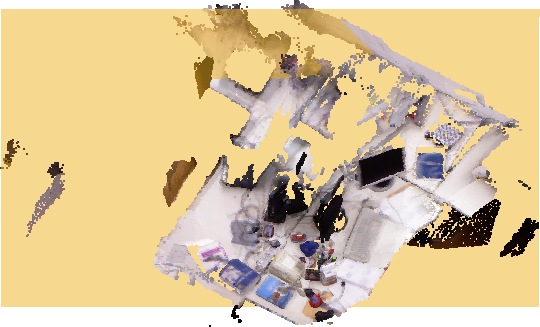}
\hfill
\includegraphics[width=0.22\linewidth, frame]{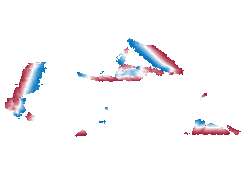}
\hfill
\includegraphics[width=0.22\linewidth, frame]{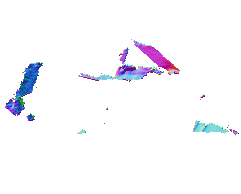}
\hfill
\includegraphics[width=0.22\linewidth, frame]{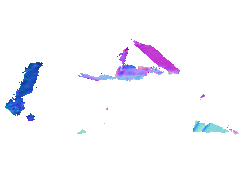}
    \\ \vspace{0.15cm}
\caption{Qualitative results for the gradient of the signed distance function of the \emph{fr1/desk1} sequence of~\cite{sturm2012benchmark}. Shown are four slices perpendicular to the $z$-axis through the SDF and the 3D gradient color-coded as normal map (after normalization), at $z=0.60$, $z=0.86$, $z=1.14$, and $z=1.90$.
The \emph{left} column shows the position of the slice in the colored 3D point cloud for better orientation.
Just like for the synthetic data in Figure~\ref{fig:sphere_gradients}, our Gradient-SDF vectors are smoother than the ones obtained using finite differences.}
\label{fig:desk_gradients}
\end{figure}

\section{Geometry after depth tracking}

\begin{figure}[H]
    \centering
    \includegraphics[height=.24\linewidth, angle=-90]{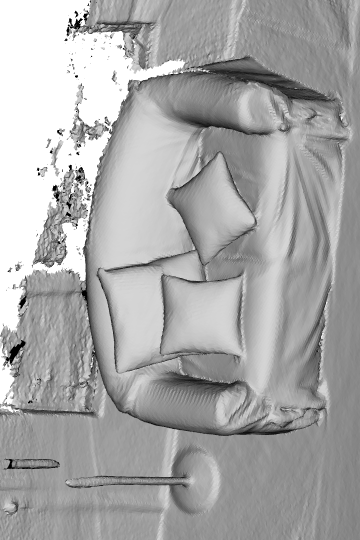}
    \hfill
    \includegraphics[height=.24\linewidth, angle=-90]{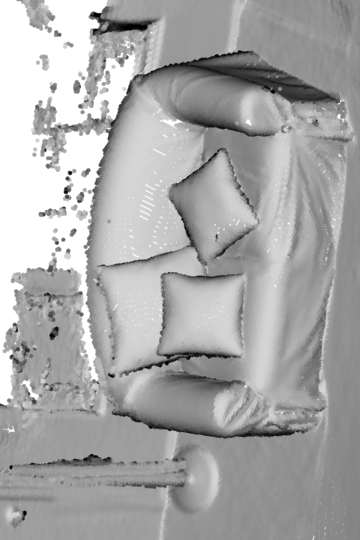}
    \hfill
    \includegraphics[height=.24\linewidth, angle=-90]{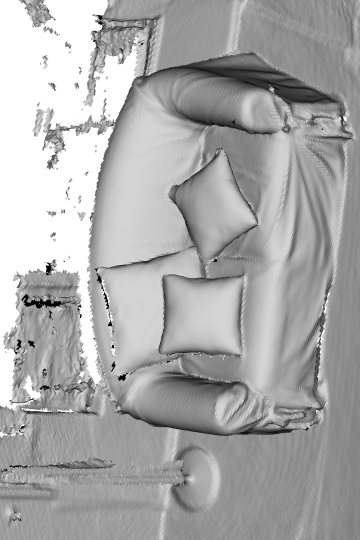}
    \hfill
    \includegraphics[height=.24\linewidth, angle=-90]{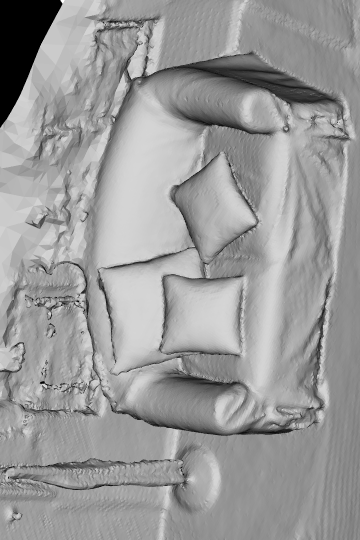}
    \vspace{1.5mm}
    \caption{Grayscale-shaded 3D reconstructions of the \emph{00577\_sofa} from~\cite{choi2016large}, see also Figure~\ref{fig:tracking_sofa}.
    \emph{Left} shows the reconstruction obtained from running SDF Tracker~\cite{canelhas2013sdf} with a $512^3$ volume.
    The remaining three figures show different visualizations of our reconstruction: an oriented point cloud (surfels), the mesh extracted using marching cubes, and the mesh reconstructed from the point cloud using SSD surface reconstruction~\cite{calakli2011ssd}.}
    \label{fig:tracking_sofa_app}
\end{figure}

\begin{figure}[H]
    \centering
    \includegraphics[height=.185\linewidth, angle=-90]{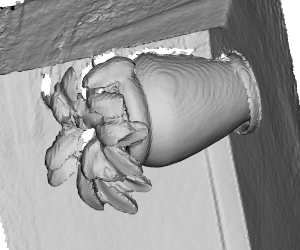}
    \hfill
    \includegraphics[height=.185\linewidth, angle=-90]{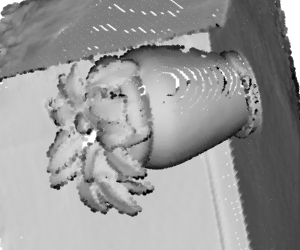}
    \hfill
    \includegraphics[height=.185\linewidth, angle=-90]{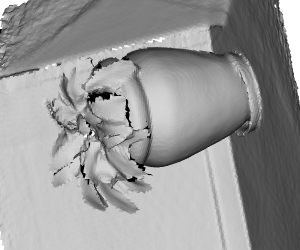}
    \hfill
    \includegraphics[height=.185\linewidth, angle=-90]{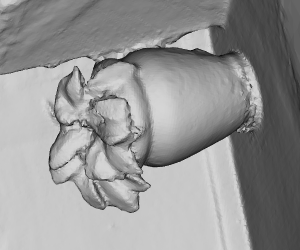}
    \vspace{1.5mm}
    \caption{Grayscale-shaded reconstructions of the \emph{06127\_plant} from~\cite{choi2016large}.
    Visualized are the same four reconstruction modes as in Figure~\ref{fig:tracking_sofa_app}, from \emph{left} to \emph{right}:
    SDF Tracker mesh, our oriented point cloud, our mesh using marching cubes, and mesh reconstruction using SSD from our point cloud.
    As explained in the main paper, our tracker can perform direct SDF tracking using a hash map, achieving comparable results to SDF Tracker while storing significantly less voxels.
    }
    \label{fig:tracking_plant}
\end{figure}

\section{Photometric Optimization}

\begin{figure}[H]
    \newcommand{\imgwidth}{0.65\linewidth}
    \newcommand{\imagewithtitle}[2]{
        \begin{tikzpicture}[baseline=0]
        \begin{scope}[local bounding box=img]
            \clip (-0.4*\imgwidth, -0.5*\imgwidth) rectangle (0.6*\imgwidth, 0.45*\imgwidth);
            \node {\includegraphics[width=\imgwidth, angle=150]{#1}};
        \end{scope}
            \node[anchor=north, align=left] at ($(img.north) + (0, -1.5)$) {\footnotesize{#2}};
        \end{tikzpicture}
    }
    
    \centering
    
\begin{minipage}{0.48\linewidth}
    \imagewithtitle{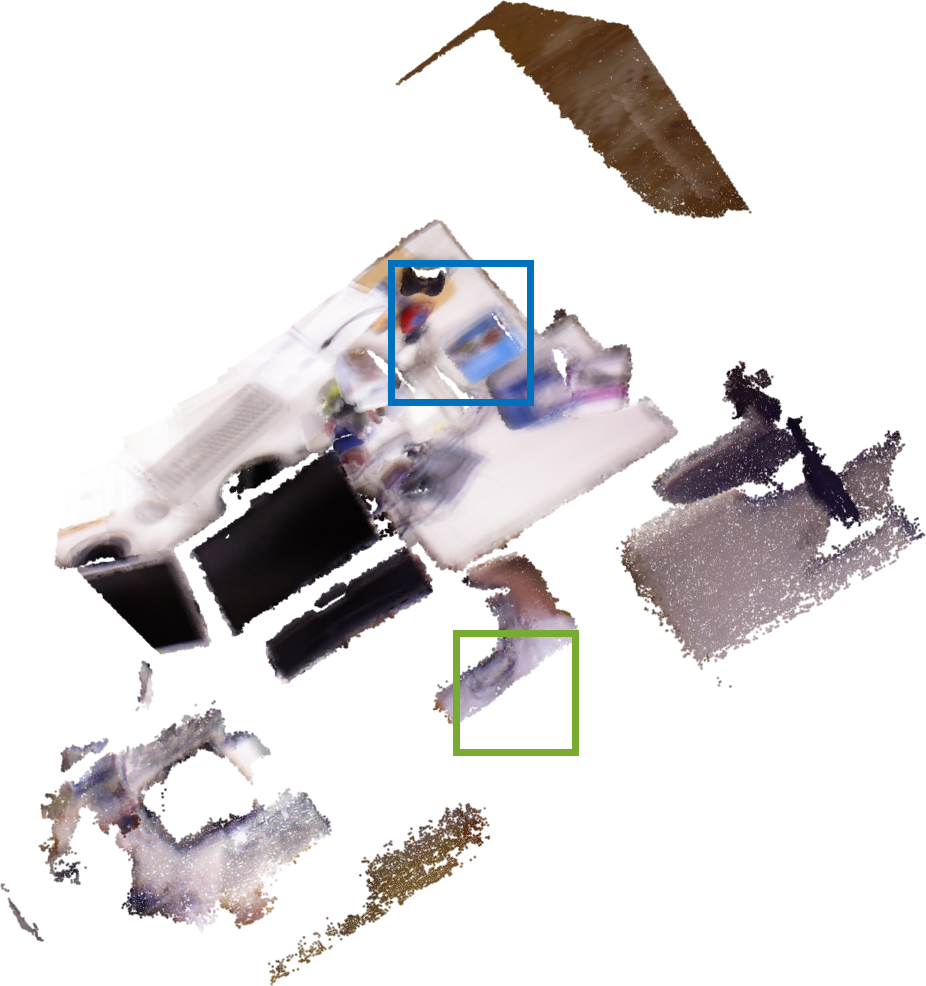}{BAD SLAM, (542K points, RMSE 1.8\,cm)}
    \begin{minipage}{0.2\linewidth}
        \centering
        \includegraphics[width=0.9\linewidth,cframe=mlgreen 1.5pt 0pt, angle=180]{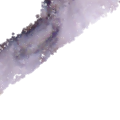}
        \includegraphics[width=0.9\linewidth,cframe=mlblue 1.5pt 0pt, angle=180]{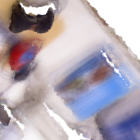}
    \end{minipage}
\end{minipage}
\begin{minipage}{0.48\linewidth}
    \imagewithtitle{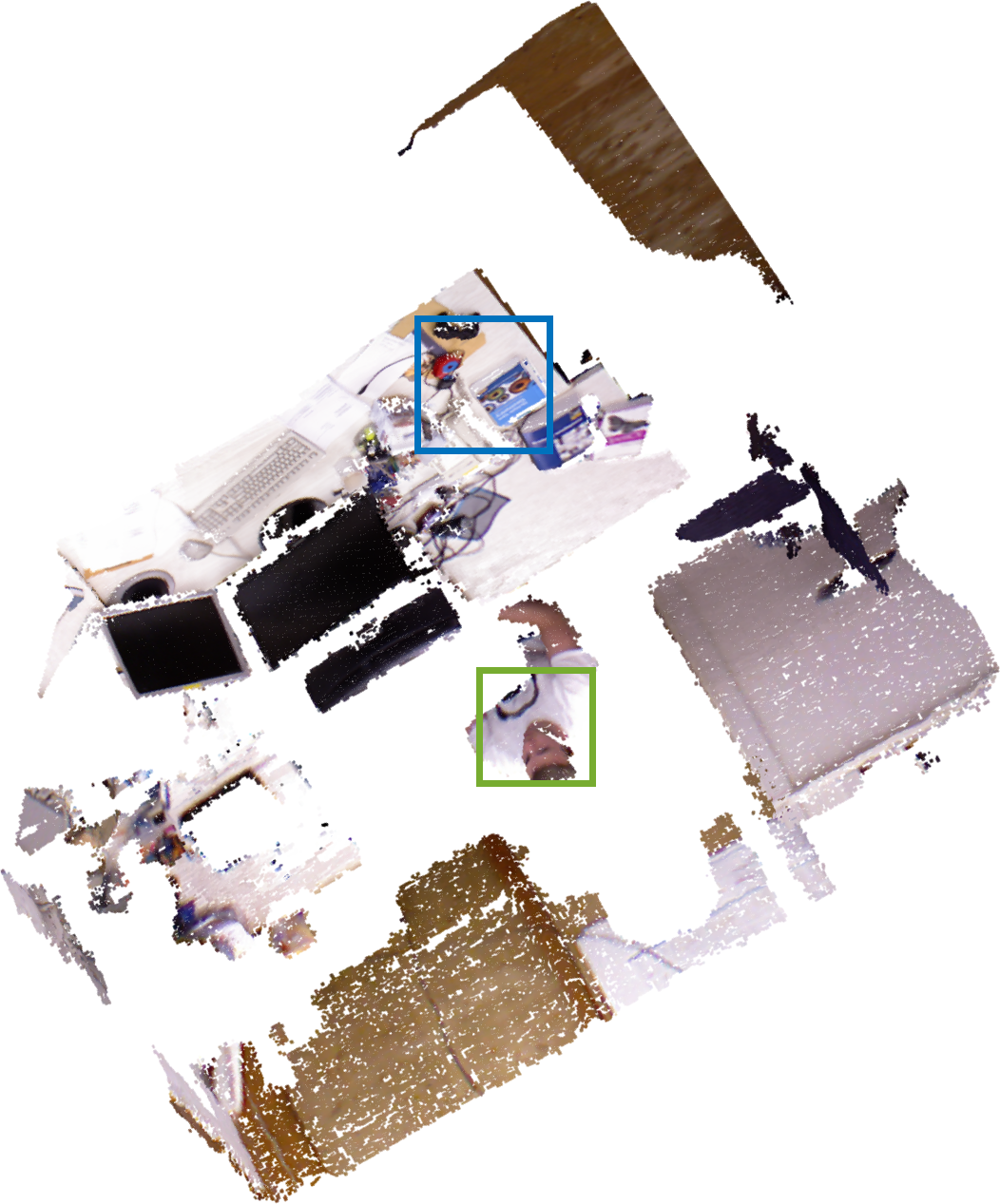}{Gradient-SDF final (373K points, RMSE 1.1\,cm)}
    \begin{minipage}{0.2\linewidth}
        \centering
        \includegraphics[width=0.9\linewidth,cframe=mlgreen 1.5pt 0pt, angle=180]{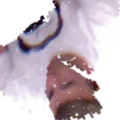}
        \includegraphics[width=0.9\linewidth,cframe=mlblue 1.5pt 0pt, angle=180]{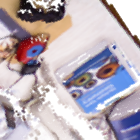}
    \end{minipage}
\end{minipage}

    \caption{Colored point cloud reconstruction produced by BAD SLAM (\emph{left}) and after optimization of the BA cost on our Gradient-SDF (\emph{right}) on the \emph{fr1/xyz} sequence of~\cite{sturm2012benchmark}.}
    \label{fig:ba_results_xyz1}
\end{figure}

\begin{figure}[H]
    \newcommand{\imgwidth}{0.75\linewidth}
    \newcommand{\imagewithtitle}[2]{
        \begin{tikzpicture}[baseline=0]
        \begin{scope}[local bounding box=img]
            \clip (-0.51*\imgwidth, -0.4*\imgwidth) rectangle (0.51*\imgwidth, 0.4*\imgwidth);
            \node {\includegraphics[width=\imgwidth, angle=180]{#1}};
        \end{scope}
            \node[anchor=north, align=left] at ($(img.north) + (0, 0.5)$) {\footnotesize{#2}};
        \end{tikzpicture}
    }
    
    \centering

\begin{minipage}{0.48\linewidth}
    \imagewithtitle{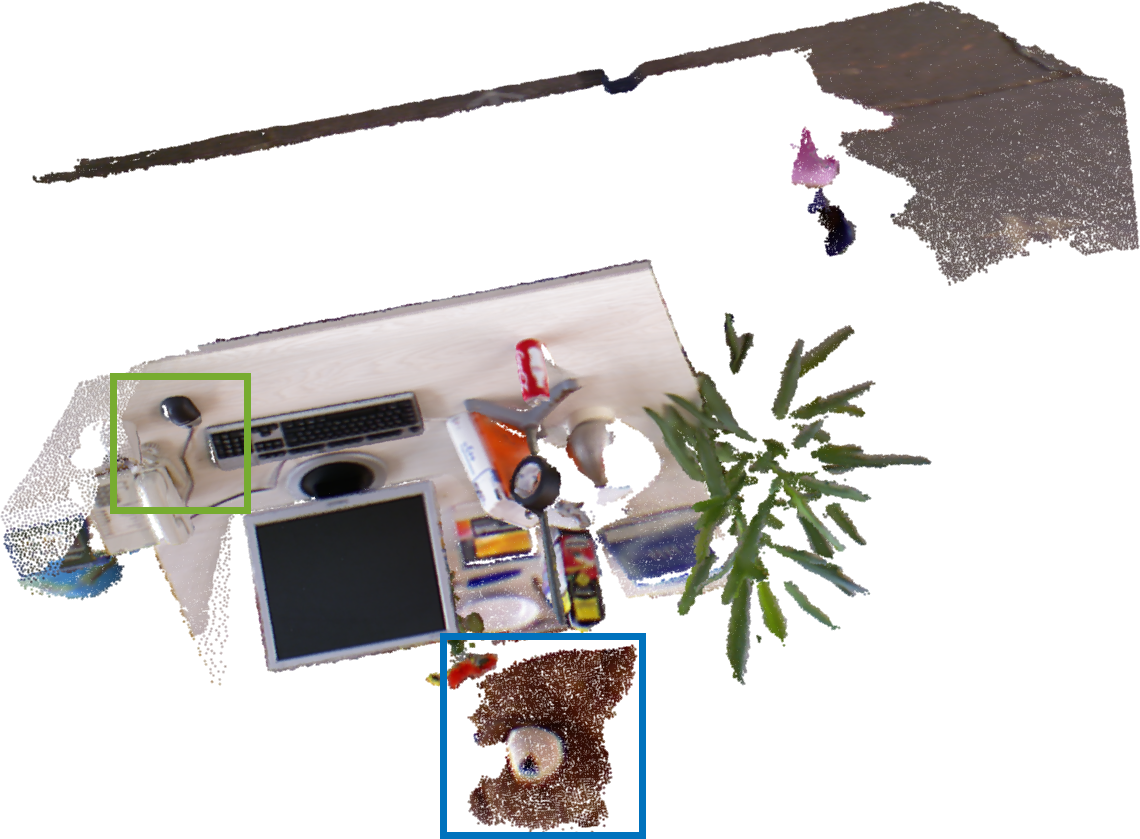}{BAD SLAM, (354K points, RMSE 1.3\,cm)}
    \begin{minipage}{0.2\linewidth}
        \centering
        \includegraphics[width=0.9\linewidth,cframe=mlgreen 1.5pt 0pt, angle=180]{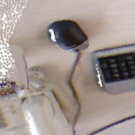}
        \includegraphics[width=0.9\linewidth,cframe=mlblue 1.5pt 0pt, angle=180]{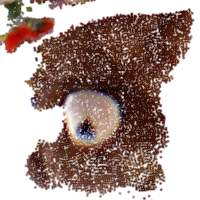}
    \end{minipage}
\end{minipage}
\begin{minipage}{0.48\linewidth}
    \imagewithtitle{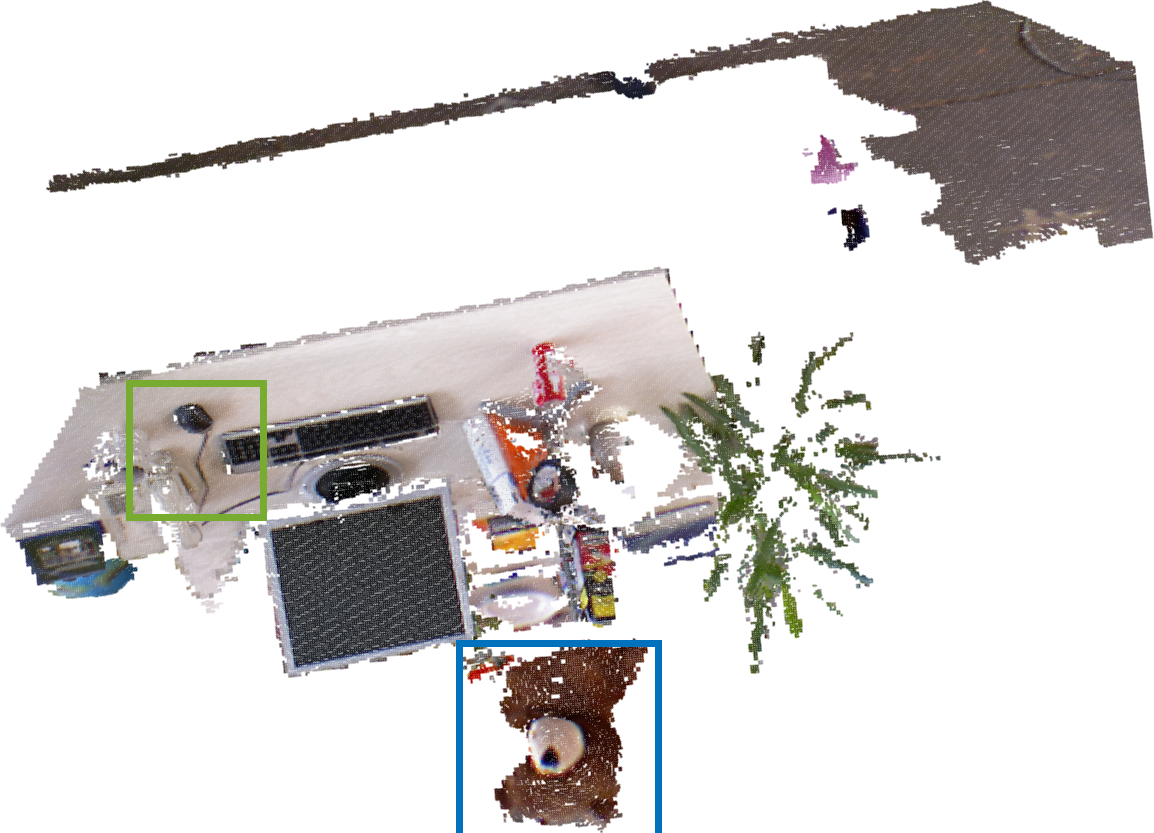}{Gradient-SDF final (201K points, RMSE 1.0\,cm)}
    \begin{minipage}{0.2\linewidth}
        \centering
        \includegraphics[width=0.9\linewidth,cframe=mlgreen 1.5pt 0pt, angle=180]{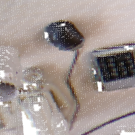}
        \includegraphics[width=0.9\linewidth,cframe=mlblue 1.5pt 0pt, angle=180]{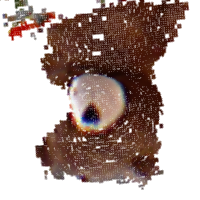}
    \end{minipage}
\end{minipage}

    \caption{Colored point cloud reconstruction produced by BAD SLAM (\emph{left}) and after optimization of the BA cost on our Gradient-SDF (\emph{right}) on the \emph{fr2/xyz} sequence of~\cite{sturm2012benchmark}.}
    \label{fig:ba_results_xyz2}
\end{figure}

\begin{figure}[H]
    \newcommand{\imgwidth}{0.75\linewidth}
    \newcommand{\imagewithtitle}[2]{
        \begin{tikzpicture}[baseline=0]
        \begin{scope}[local bounding box=img]
            \clip (-0.51*\imgwidth, -0.4*\imgwidth) rectangle (0.51*\imgwidth, 0.4*\imgwidth);
            \node {\includegraphics[width=\imgwidth]{#1}};
        \end{scope}
            \node[anchor=north, align=left] at ($(img.north) + (0, 0.5)$) {\footnotesize{#2}};
        \end{tikzpicture}
    }
    
    \centering

\begin{minipage}{0.48\linewidth}
    \imagewithtitle{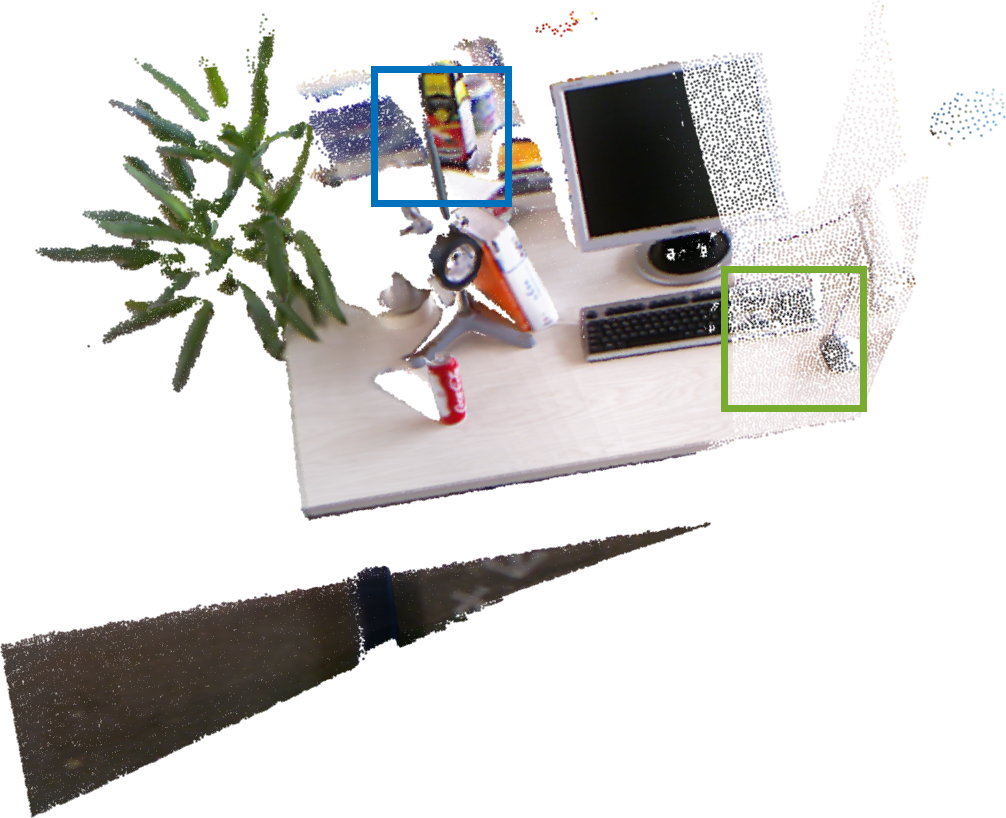}{BAD SLAM, (383K points, RMSE 0.9\,cm)}
    \begin{minipage}{0.2\linewidth}
        \centering
        \includegraphics[width=0.9\linewidth,cframe=mlgreen 1.5pt 0pt]{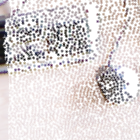}
        \includegraphics[width=0.9\linewidth,cframe=mlblue 1.5pt 0pt]{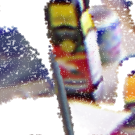}
    \end{minipage}
\end{minipage}
\begin{minipage}{0.48\linewidth}
    \imagewithtitle{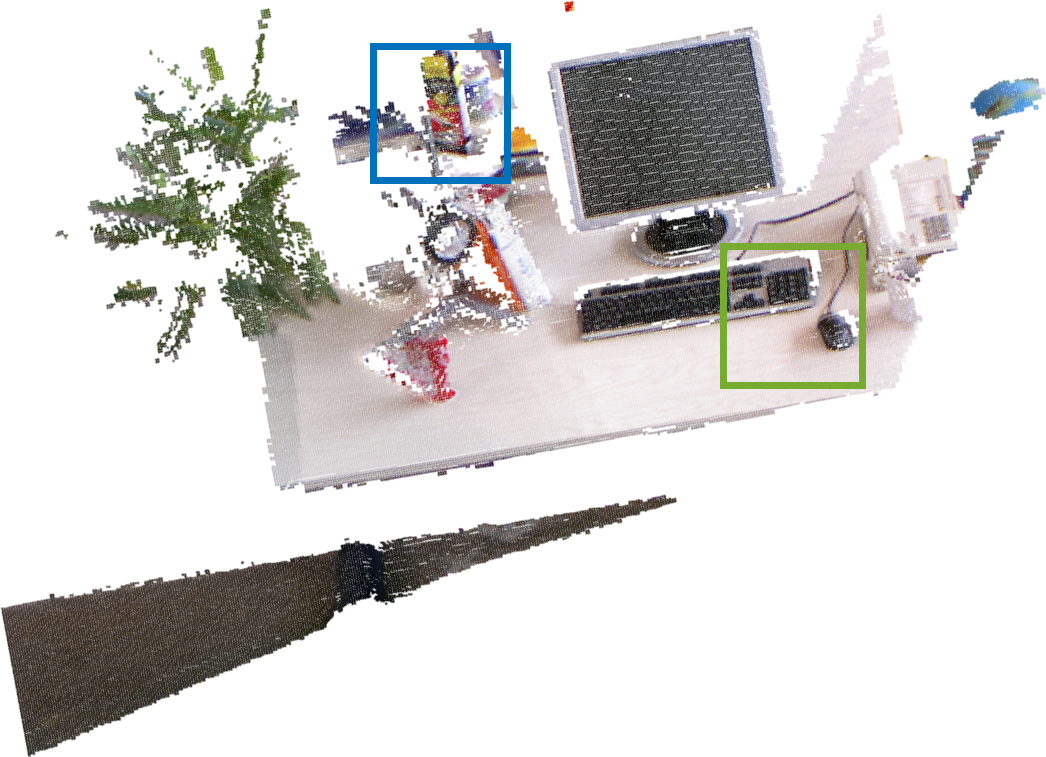}{Gradient-SDF final (338K points, RMSE 2.4\,cm)}
    \begin{minipage}{0.2\linewidth}
        \centering
        \includegraphics[width=0.9\linewidth,cframe=mlgreen 1.5pt 0pt]{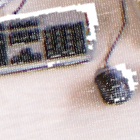}
        \includegraphics[width=0.9\linewidth,cframe=mlblue 1.5pt 0pt]{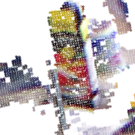}
    \end{minipage}
\end{minipage}

    \caption{Colored point cloud reconstruction produced by BAD SLAM (\emph{left}) and after optimization of the BA cost on our Gradient-SDF (\emph{right}) on the \emph{fr2/rpy} sequence of~\cite{sturm2012benchmark}.}
    \label{fig:ba_results_rpy2}
\end{figure}

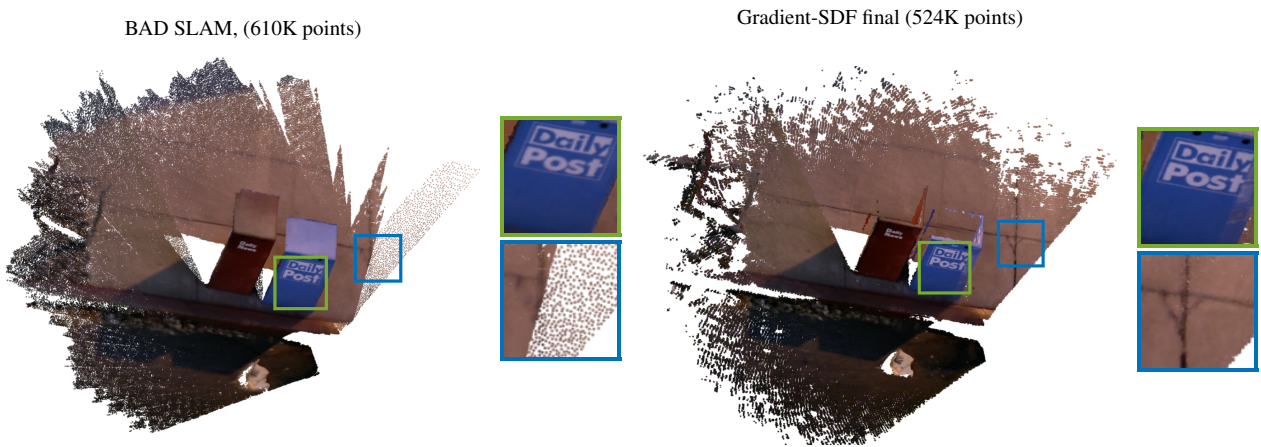
\begin{figure}[H]
    \newcommand{\imgwidth}{0.75\linewidth}
    \newcommand{\imagewithtitle}[2]{
        \begin{tikzpicture}[baseline=0]
        \begin{scope}[local bounding box=img]
            \clip (-0.51*\imgwidth, -0.4*\imgwidth) rectangle (0.51*\imgwidth, 0.4*\imgwidth);
            \node {\includegraphics[width=\imgwidth]{#1}};
        \end{scope}
            \node[anchor=north, align=left] at ($(img.north) + (0, 0.5)$) {\footnotesize{#2}};
        \end{tikzpicture}
    }
    
    \centering

\begin{minipage}{0.48\linewidth}
    \imagewithtitle{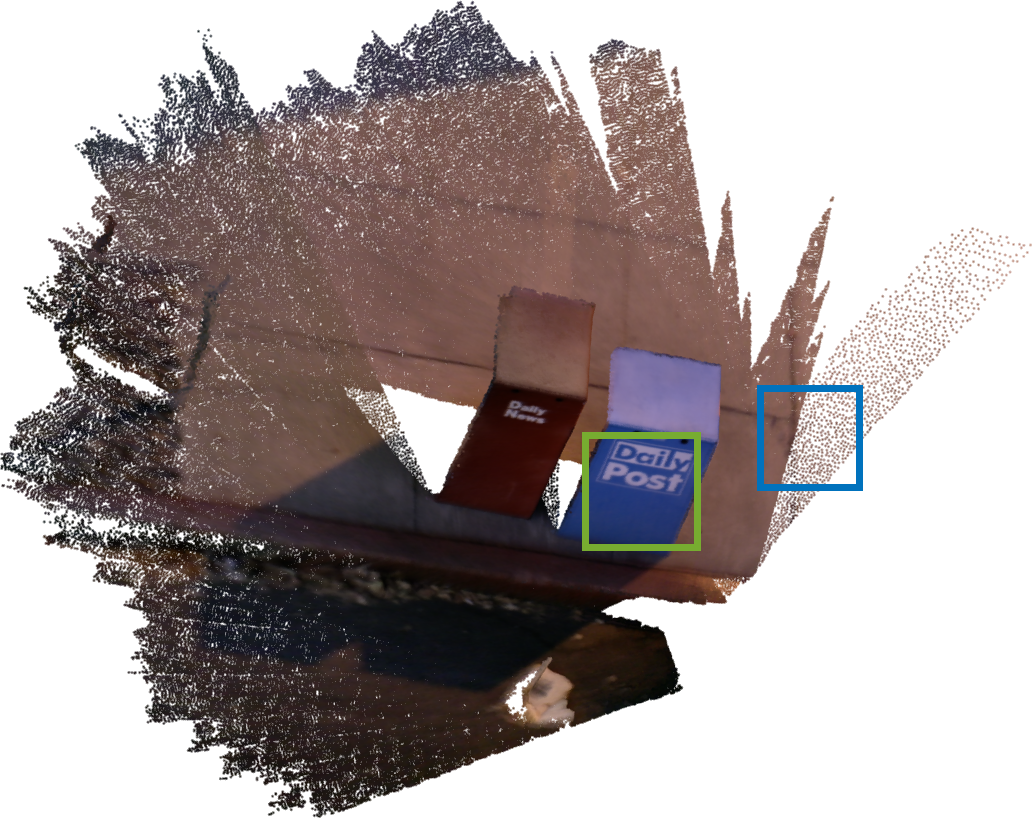}{BAD SLAM, (610K points)}
    \begin{minipage}{0.2\linewidth}
        \centering
        \includegraphics[width=0.9\linewidth,cframe=mlgreen 1.5pt 0pt]{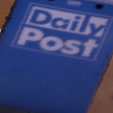}
        \includegraphics[width=0.9\linewidth,cframe=mlblue 1.5pt 0pt]{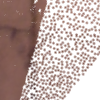}
    \end{minipage}
\end{minipage}
\begin{minipage}{0.48\linewidth}
    \imagewithtitle{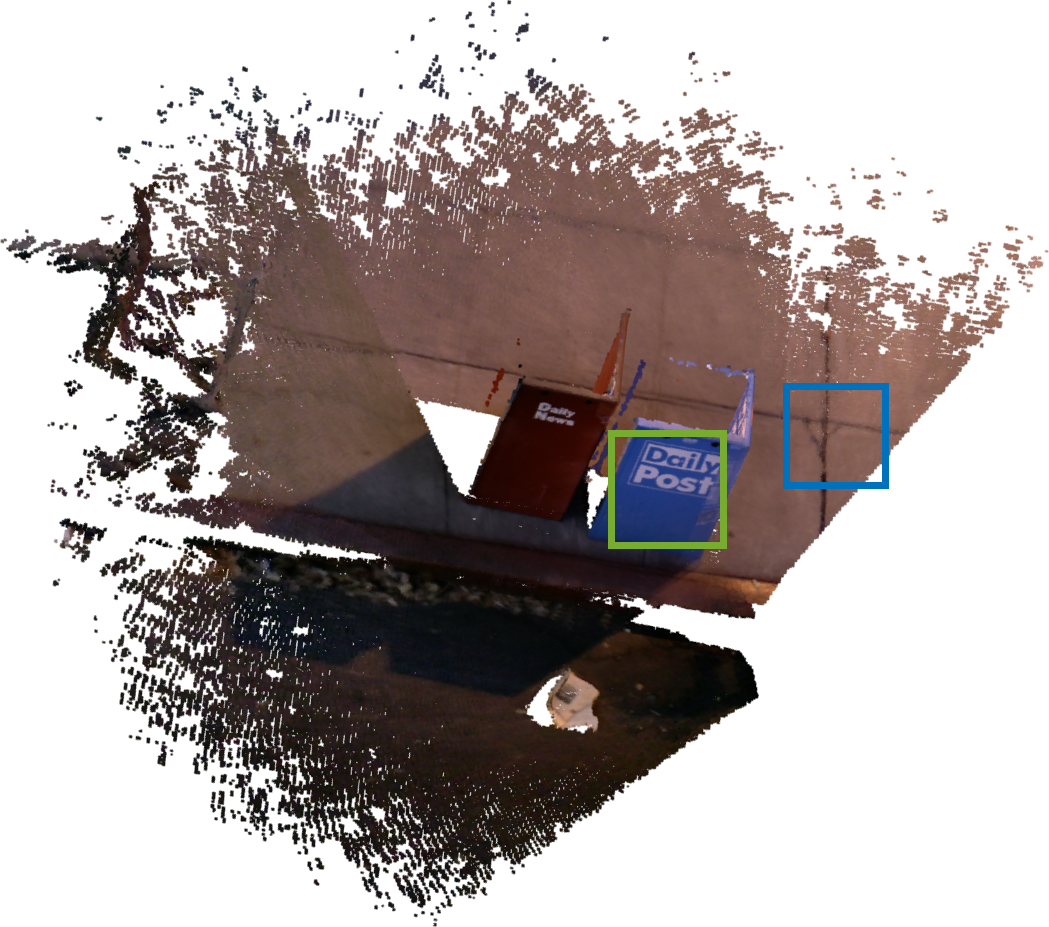}{Gradient-SDF final (524K points)}
    \begin{minipage}{0.2\linewidth}
        \centering
        \includegraphics[width=0.9\linewidth,cframe=mlgreen 1.5pt 0pt]{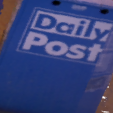}
        \includegraphics[width=0.9\linewidth,cframe=mlblue 1.5pt 0pt]{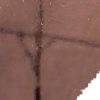}
    \end{minipage}
\end{minipage}

\caption{Colored point cloud reconstruction produced by BAD SLAM~\cite{schops2019bad} (\emph{left}) and after optimization of the BA cost on our Gradient-SDF (\emph{right}) on the \emph{01847\_kiosk} sequence of~\cite{choi2016large}. For this dataset, no ground truth poses are available, thus no ATE is stated.}
\label{fig:ba_results_kiosk}
\end{figure}

\begin{figure}[H]
    \newcommand{\imgwidth}{0.6\linewidth}
    \newcommand{\imagewithtitle}[2]{
        \begin{tikzpicture}[baseline=0]
        \begin{scope}[local bounding box=img]
            \clip (-0.61*\imgwidth, -0.51*\imgwidth) rectangle (0.61*\imgwidth, 0.51*\imgwidth);
            \node {\includegraphics[width=.4\linewidth, angle=270]{#1}};
        \end{scope}
            \node[anchor=north, align=left] at ($(img.north) + (0, 0)$) {\footnotesize{#2}};
        \end{tikzpicture}
    }
    
    \centering

\begin{minipage}{0.48\linewidth}
    \imagewithtitle{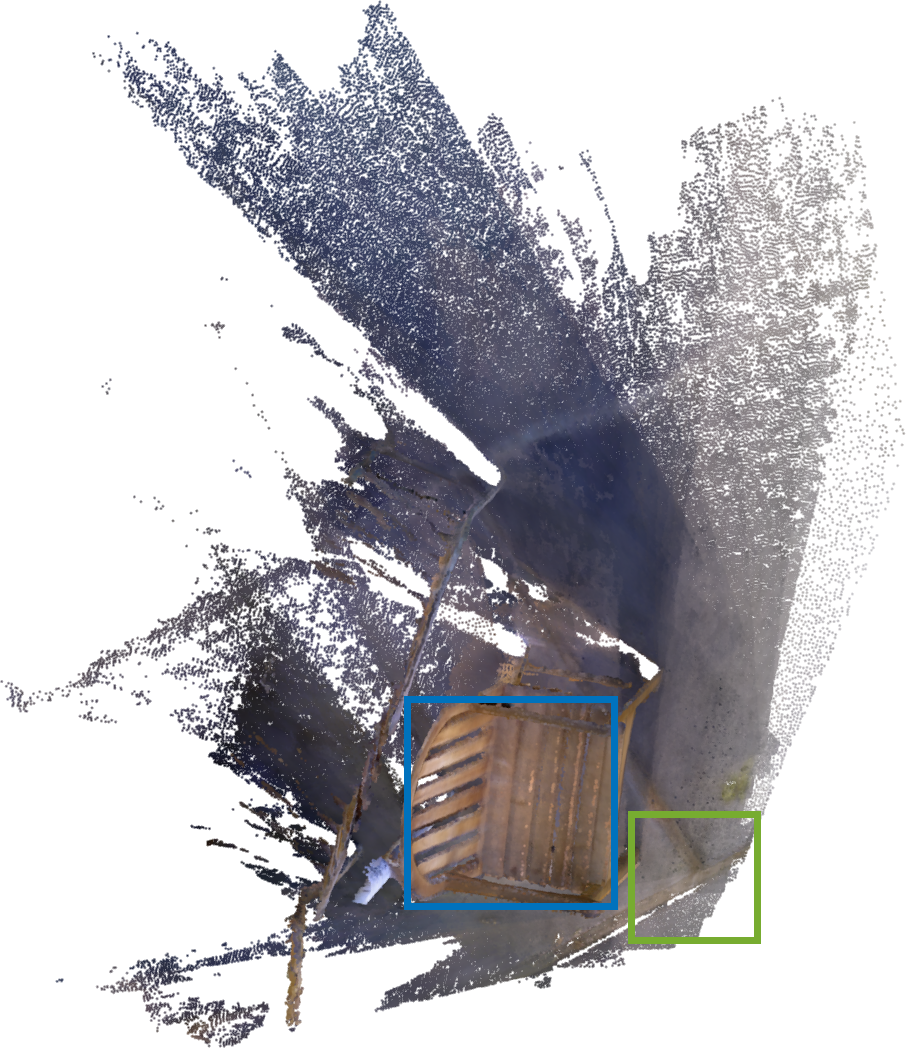}{BAD SLAM, (742K points)}
    \begin{minipage}{0.2\linewidth}
        \centering
        \includegraphics[width=0.9\linewidth,cframe=mlgreen 1.5pt 0pt, angle=270]{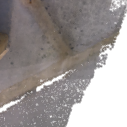}
        \includegraphics[width=0.9\linewidth,cframe=mlblue 1.5pt 0pt, angle=270]{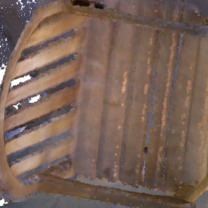}
    \end{minipage}
\end{minipage}
\begin{minipage}{0.48\linewidth}
    \imagewithtitle{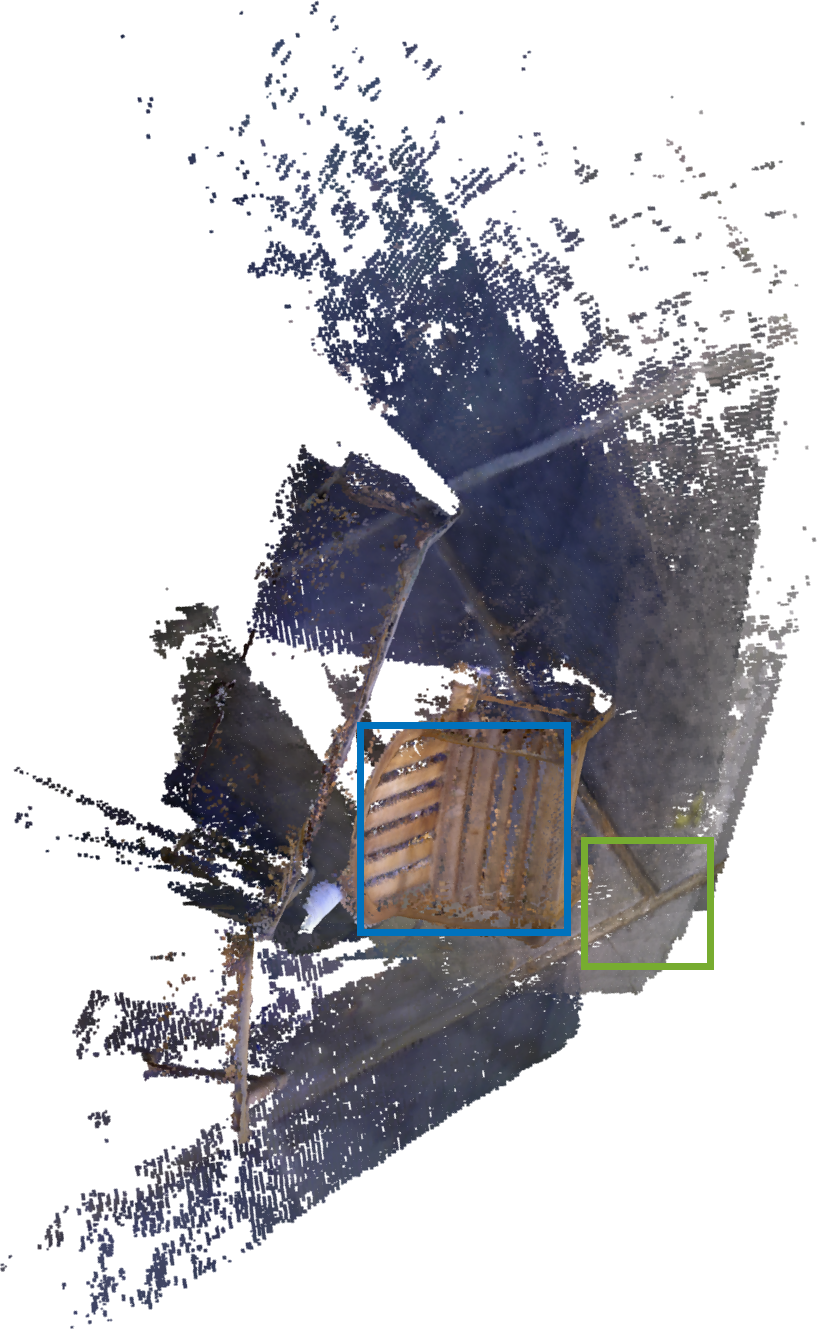}{Gradient-SDF final (341K points)}
    \begin{minipage}{0.2\linewidth}
        \centering
        \includegraphics[width=0.9\linewidth,cframe=mlgreen 1.5pt 0pt, angle=270]{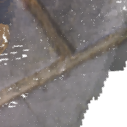}
        \includegraphics[width=0.9\linewidth,cframe=mlblue 1.5pt 0pt, angle=270]{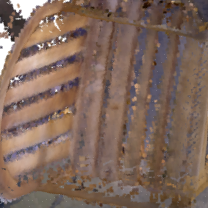}
    \end{minipage}
\end{minipage}
\vspace{-1cm}
\caption{Colored point cloud reconstruction produced by BAD SLAM~\cite{schops2019bad} (\emph{left}) and after optimization of the BA cost on our Gradient-SDF (\emph{right}) on the \emph{08309\_woodenchair} sequence of~\cite{choi2016large}.}
\label{fig:ba_results_woodenchair}
\end{figure}
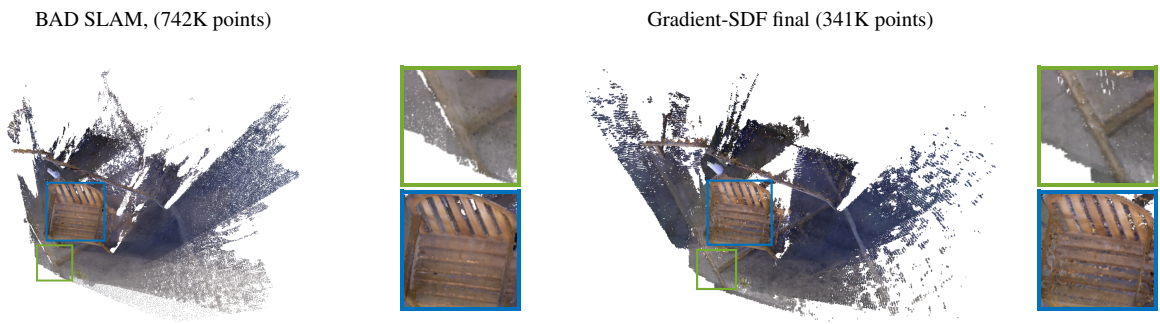

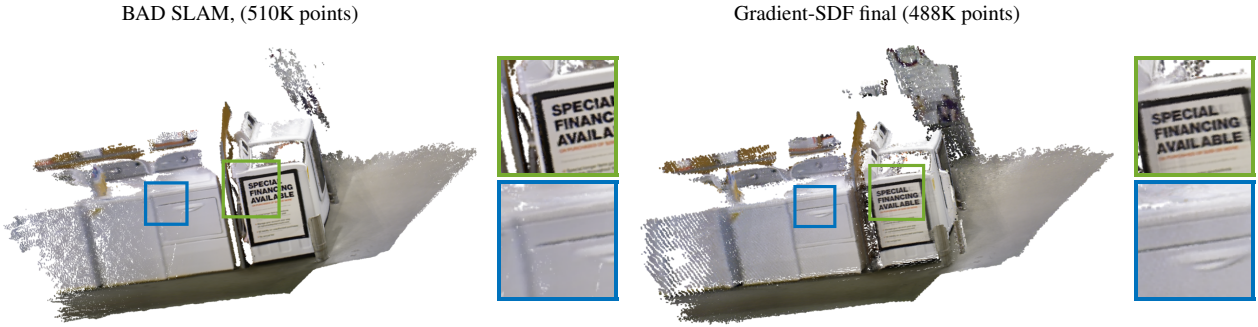
\begin{figure}[H]
    \newcommand{\imgwidth}{0.75\linewidth}
    \newcommand{\imagewithtitle}[2]{
        \begin{tikzpicture}[baseline=0]
        \begin{scope}[local bounding box=img]
            \clip (-0.51*\imgwidth, -0.4*\imgwidth) rectangle (0.51*\imgwidth, 0.4*\imgwidth);
            \node {\includegraphics[width=\imgwidth, angle=180]{#1}};
        \end{scope}
            \node[anchor=north, align=left] at ($(img.north) + (0, 0)$) {\footnotesize{#2}};
        \end{tikzpicture}
    }
    
    \centering

\begin{minipage}{0.48\linewidth}
    \imagewithtitle{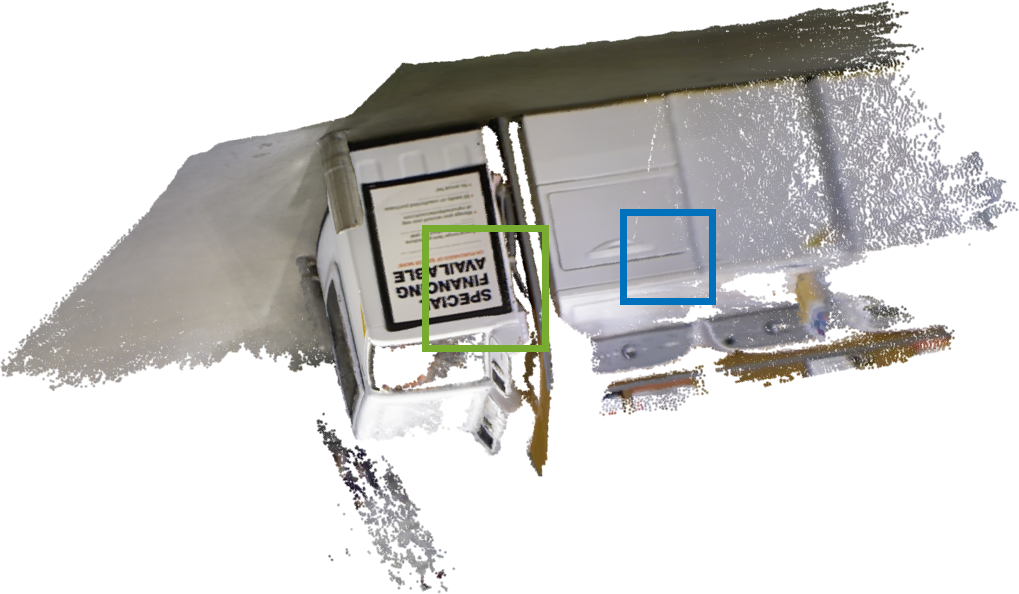}{BAD SLAM, (510K points)}
    \begin{minipage}{0.2\linewidth}
        \centering
        \includegraphics[width=0.9\linewidth,cframe=mlgreen 1.5pt 0pt, angle=180]{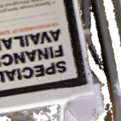}
        \includegraphics[width=0.9\linewidth,cframe=mlblue 1.5pt 0pt, angle=180]{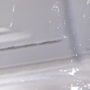}
    \end{minipage}
\end{minipage}
\begin{minipage}{0.48\linewidth}
    \imagewithtitle{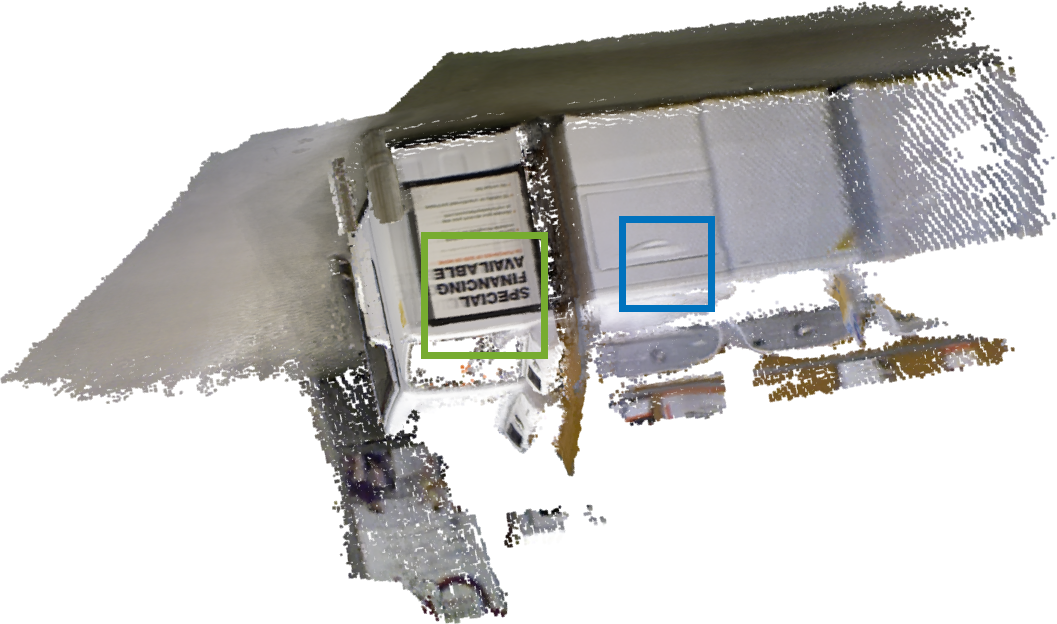}{Gradient-SDF final (488K points)}
    \begin{minipage}{0.2\linewidth}
        \centering
        \includegraphics[width=0.9\linewidth,cframe=mlgreen 1.5pt 0pt, angle=180]{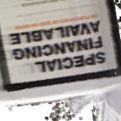}
        \includegraphics[width=0.9\linewidth,cframe=mlblue 1.5pt 0pt, angle=180]{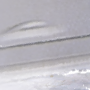}
    \end{minipage}
\end{minipage}
\caption{Colored point cloud reconstruction produced by BAD SLAM~\cite{schops2019bad} (\emph{left}) and after optimization of the BA cost on our Gradient-SDF (\emph{right}) on the \emph{05635\_washmachine} sequence of~\cite{choi2016large}.}
\label{fig:ba_results_washmachine}
\end{figure}

\section{Used code and data}

\begin{table}[H]
    \centering
    \begin{longtable}{l p{4.25cm} l l p{6cm} l}
    \toprule
         && code/data & year & link & license \\
    \midrule
        \cite{sturm2012benchmark} & TUM RGB-D Benchmark & dataset & 2012 &  {\small \url{https://vision.in.tum.de/data/datasets/rgbd-dataset}} &  CC BY 4.0 \\
        \cite{choi2016large} & Redwood Large Dataset of Object Scans & dataset & 2016 & {\small\url{http://www.redwood-data.org/3dscan/}} & Public Domain \\
        \cite{calakli2011ssd} & SSD Surface Reconstruction (version 3.0) & code & 2011 & {\small\url{http://mesh.brown.edu/ssd/software.html}} & BSD-3 \\
        \cite{sturm2012benchmark} & TUM RGB-D Benchmark & code & 2012 & {\small\url{https://vision.in.tum.de/data/datasets/rgbd-dataset/tools}} & BSD-2 \\
        \cite{canelhas2013sdf} & SDF Tracker & code & 2013 & {\small\url{https://wiki.ros.org/sdf_tracker}} & BSD-3 \\
        \cite{schops2019bad} & BAD SLAM & code & 2019 & {\small\url{https://github.com/ETH3D/badslam}} & BSD-3 \\
    \bottomrule
    \caption{Datasets and code used in this work, together with reference, link and license. We did our real-world experiments on two datasets, TUM RGB-D (for which ground truth poses exist), and Redwood LOD (without ground truth). For code, we used the smooth signed distance surface reconstruction (SSD) to generate a mesh from our output surfel cloud, and SDF Tracker to reconstruct geometry following~\cite{canelhas2013sdf}, see Figures~\ref{fig:tracking_sofa_app} and \ref{fig:tracking_plant}. The TUM RGB-D code was used to compute ATEs of estimated camera poses, and BAD SLAM for qualitative and quantitative results of dense RGB-D bundle adjustment.}
    \label{tab:code_data}
    \end{longtable}
    \vspace{0.25cm}
\end{table}

\end{document}